\documentclass[letterpaper,journal]{IEEEtran}

\IEEEoverridecommandlockouts                              

\usepackage{amsthm}
\usepackage{graphicx}
\usepackage{caption}
\usepackage{subcaption}
\usepackage[linesnumbered,ruled,vlined]{algorithm2e}
\usepackage{algorithmicx}
\usepackage{amsmath}
\theoremstyle{plain}
\usepackage{mathtools}
\usepackage{epsfig}
\usepackage{amssymb}
\usepackage{epstopdf}
\usepackage{multicol}
\usepackage{multirow}
\usepackage{array}
\usepackage{url}
\usepackage{color}
\usepackage{float}
\usepackage{wrapfig}
\usepackage{placeins}

\def\BibTeX{{\rm B\kern-.05em{\sc i\kern-.025em b}\kern-.08em
    T\kern-.1667em\lower.7ex\hbox{E}\kern-.125emX}}

\usepackage{blindtext}
\usepackage{nidanfloat}

\title{\LARGE \bf
Motion Planning Networks: Bridging the Gap Between Learning-based and Classical Motion Planners 
}

\author{Ahmed H. Qureshi, Yinglong Miao, Anthony Simeonov and Michael C. Yip
\thanks{A. H. Qureshi, Y. Miao and M. Yip are affiliated with University of California San Diego, La Jolla, CA, USA. {\tt\small \{a1qureshi, y2miao, yip\}@ucsd.edu}. A. Simeonov is affiliated with Massachusetts Institute of Technology, Cambridge, MA, USA. {\tt\small asimeono@mit.edu}}
\thanks{This work has been accepted for publication at IEEE Transactions on Robotics.}
}

\begin{document}
%

\maketitle
\thispagestyle{empty}
\pagestyle{empty}


\begin{abstract}
This paper describes Motion Planning Networks (MPNet)\footnote{Supplementary material including implementation parameters and project videos are available at https://sites.google.com/view/mpnet/home.}, a computationally efficient, learning-based neural planner for solving motion planning problems. MPNet uses neural networks to learn general near-optimal heuristics for path planning in seen and unseen environments. It takes environment information such as raw point-cloud from depth sensors, as well as a robot's initial and desired goal configurations and recursively calls itself to bidirectionally generate connectable paths. In addition to finding directly connectable and near-optimal paths in a single pass, we show that worst-case theoretical guarantees can be proven if we merge this neural network strategy with classical sample-based planners in a hybrid approach while still retaining significant computational and optimality improvements. To train the MPNet models, we present an active continual learning approach that enables MPNet to learn from streaming data and actively ask for expert demonstrations when needed, drastically reducing data for training. We validate MPNet against gold-standard and state-of-the-art planning methods in a variety of problems from 2D to 7D robot configuration spaces in challenging and cluttered environments, with results showing significant and consistently stronger performance metrics, and motivating neural planning in general as a modern strategy for solving motion planning problems efficiently.

\end{abstract}
\section{Introduction}
\begin{figure}\vspace*{0.1in}
    \centering
   \begin{subfigure}[b]{0.23\textwidth}
       \includegraphics[height=3.4cm]{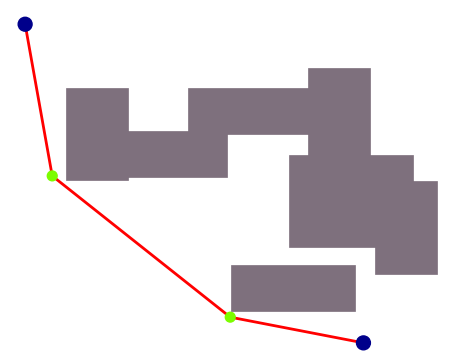}
       \caption{MPNet}
    \end{subfigure}
     \begin{subfigure}[b]{0.23\textwidth}
        \includegraphics[height=3.4cm]{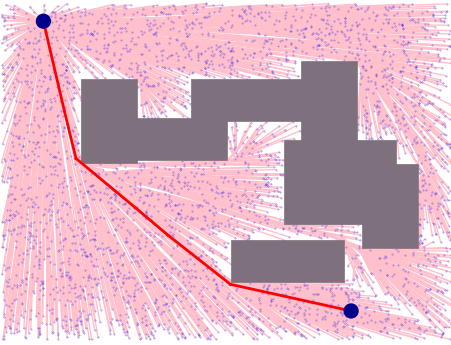}
        \caption{RRT*}
    \end{subfigure}
         \begin{subfigure}[b]{0.23\textwidth}
        \includegraphics[height=3.4cm]{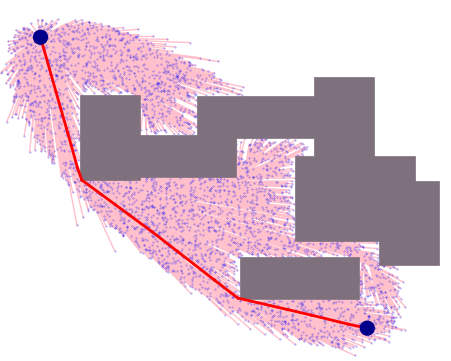}
        \caption{Informed-RRT*}
    \end{subfigure}
     \begin{subfigure}[b]{0.23\textwidth}
        \includegraphics[height=3.4cm]{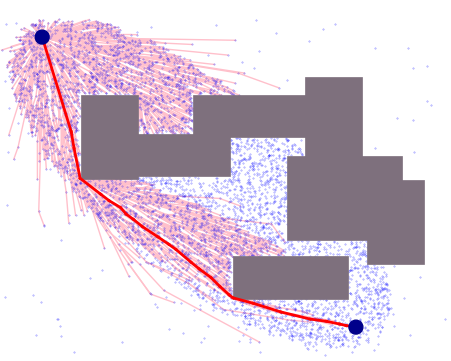}
        \caption{BIT*}
    \end{subfigure}
    \caption{MPNet can greedily lay out a near-optimal path after having past experiences in similar environments, whereas classical planning methods such as RRT* \cite{karaman2011sampling}, Informed-RRT* \cite{gammell2014informed}, and BIT* \cite{gammell2015batch} need to expand their planning spaces through the exhaustive search before finding a similarly optimal path.}\label{mpnet0}
\vspace{-0.1in}\end{figure}
Motion planning is among the core research problems in robotics and artificial intelligence. It aims to find a collision-free, low-cost path connecting a start and goal configuration for an agent \cite{lavalle2006planning} \cite{latombe2012robot}. An ideal motion planning algorithm for solving real-world problems should offer following key features: 
(i) completeness and optimality guarantees --- implying that a solution will be found if one exists and that the solution will be globally optimal,
(ii) computational efficiency --- finding a solution in either real-time or in sub-second times or better while being memory efficient, and
(iii) insensitivity to environment complexity --- the algorithm is effective and efficient in finding solutions regardless of the constraints of the environment.
Decades of research have produced many significant milestones for motion planning include resolution-complete planners such artificial potential fields \cite{khatib1986real}, sample-based motion planners such as Rapidly-Exploring Random Trees (RRT) \cite{lavalle2006planning} and its optimal variant RRT* \cite{karaman2011sampling}, heuristically biased solvers such as \cite{gammell2015batch} and lazy search methods \cite{haghtalab2018provable}. However, each planner and their variants have tradeoffs amongst the ideal features of motion planners. Thus, no single motion planner has emerged above all others to solve a broad range of problems.
 
A recent research wave has led to the cross-fertilization of motion planning and machine learning to solve planning problems. Motion planning and machine learning for control are both well established and active research areas where huge progress has been made. In our initial work in motion planning with neural networks \cite{qureshi2018motion} \cite{qureshi2018deeply}, we highlighted that merging both fields holds a great potential to build motion planning methods with all key features of an ideal planner ranging from theoretical guarantees to computational efficiency. In this paper, we formally present Motion Planning Networks, or MPNet, its features corresponding to an ideal planner and its merits in solving complex robotic motion planning problems. 

MPNet is a deep neural network-based bidirectional iterative planning algorithm that comprises two modules, an encoder network and planning network. The encoder network takes the environment information, such as the raw or voxelized output of a depth camera or LIDAR, and embeds them into a latent space. The planning network takes the environment encoding, robot's current and goal state, and outputs a next state of the robot that would lead it closer to the goal region. MPNet can very effectively generate steps from start to goal that are likely to be part of the optimal solution with minimal-to-no branching required. Being a neural network approach, we also propose three learning strategies to train MPNet: i) offline batch learning which assumes the availability of all training data, ii) continual learning with episodic memory which assumes that the expert demonstrations come in streams and the global training data distribution is unknown, and iii) active continual learning that incorporates MPNet into the learning process and asks for expert demonstrations only when needed. 
The following are the major contributions of MPNet:
\begin{itemize}
\item MPNet can learn from streaming data which is crucial for real-world scenarios in which the expert demonstrations usually come in streams, such as in semi self-driving cars. However, as MPNet uses deep neural networks, it can suffer from catastrophic forgetting when given the data in streams. To retain MPNet prior knowledge, we use a continual learning approach based on episodic memory and constraint optimization.
\item The active continual learning approach that asks for demonstrations only when needed, hence improving the overall training data efficiency. This strategy is in response to practical and data-efficient learning where planning problems come in streams, and MPNet attempts to plan a motion for them. In case MPNet fails to find a path for a given problem, only then an Oracle Planner is called to provide an expert demonstration for learning.
\item MPNet plans paths with a low computational complexity and exhibits a mean computation time of less than 1 second in all presented experiments.
\item MPNet can generate informed samples, thanks to its stochastic planning network, for sampling-based motion planners such as RRT* without incurring any additional computational load. The MPNet informed sampling based RRT* exhibits mean computation time of less than a second while ensuring asymptotic optimality and completeness guarantees.

\item A hybrid planning approach that combines MPNet with classical planners to provide worst-case guarantees of our approach. MPNet plans motion through divide-and-conquer since it first outputs a set of critical states and recursively finds paths between them. Therefore, it is straightforward to outsource a segment of a planning problem to a classical planner, if needed, while retaining the computational benefits of MPNet. 
\item MPNet generalizes to similar but unseen environments that were not in the training examples.
\end{itemize}  

We organize the remaining paper as follows. Section II provides a thorough literature review of classical and learning-based planning methods. Section III describes notations required to outline MPNet, and our propositions highlighting the key features of our method. Section IV presents approaches to train the neural models whereas Section V outlines our novel neural-network-based iterative planning algorithms. Section VI presents results followed by  Section VII that provides discussion and theoretical analysis concerning worst-cases guarantees for MPNet. Section VIII concludes the paper with consideration of future avenues of development. 

\section{Related Work} 

The quest for solving the motion planning problem originated with the development of complete algorithms which suffer from computational inefficiency, and thus led to more efficient  methods with resolution- and probabilistic- completeness \cite{lavalle2006planning}. The algorithms with full completeness guarantees \cite{schwartz1983piano} \cite{lozano1979algorithm} find a path solution, if one exists, in a finite-time. However, these methods are computationally intractable for most practical applications as they require the complete environment information such as obstacle geometry which is not usually available in real-world scenarios \cite{canny1988complexity}. The algorithms with resolution-completeness \cite{khatib1986real} \cite{brooks1985subdivision} also find a path, if one exists, in a finite-time, but require tedious fine-tuning of resolution parameters for every given planning problem. To overcome the limitations of complete and resolution-complete algorithms, the probabilistically complete methods, also known as Sampling-based Motion Planners (SMPs), were introduced \cite{lavalle2006planning}. These methods rely on sampling techniques to generate rapidly-exploring trees or roadmaps in robot's obstacle-free state-space. The feasible path is determined by querying the generated graph through shortest path-finding methods such as Dijkstra's method. These methods are probabilistically complete, since the probability of finding a path solution, if one exists, approaches one as the number of samples in the graph approaches infinity \cite{lavalle2006planning}.

The prominent and widely used SMPs include single-query rapidly-exploring random trees (RRT) \cite{lavalle1998rapidly} and multi-query probabilistic roadmaps (PRM) \cite{kavraki1998probabilistic}. In practice, single-query methods are preferred since most multi-query problems can be solved as a series of single-query problems \cite{lavalle1998rapidly} \cite{karaman2011sampling}. Besides, PRM based methods require pre-computation of a roadmap which is usually expensive to determine in online planning problems \cite{karaman2011sampling}. Therefore, RRT and its variants have now emerged as a promising tools for motion planning that finds a path irrespective of obstacles geometry. Although the RRT algorithm rapidly finds a path solution, if one exists, they fail to find the shortest path solution \cite{karaman2011sampling}. An optimal variant of RRT called RRT* asymptotically guarantees to find the shortest path, if one exists \cite{karaman2011sampling}. However, RRT* becomes computationally inefficient as the dimensionality of the planning problem increases. Furthermore, studies show that to determine a $\epsilon$-near optimal path in $d \in \mathbb{N}$ dimensions, nearly $O(1/\epsilon^d)$ samples are required. Thus, RRT* is no better than grid search methods in higher dimensional spaces \cite{hauser2015lazy}. 
Several methods have been proposed to mitigate limitations in current asymptotically optimal SMPs through different heuristics such as biased sampling \cite{qureshi2016potential} \cite{qureshi2013adaptive}
\cite{gammell2014informed} \cite{gammell2015batch}, lazy edge evaluation \cite{haghtalab2018provable}, and bidirectional tree generations \cite{qureshi2015intelligent} \cite{tahir2018potentially}. 

Biased sampling heuristics adaptively sample the robot state-space to overcome limitations caused by random uniform exploration in underlying SMP methods. For instance, P-RRT* \cite{qureshi2016potential} \cite{qureshi2013adaptive} incorporates artificial potential fields \cite{khatib1986real} into RRT* to generate goal directed trees for rapid convergence to an optimal path solution. In similar vein, Gammell et al. proposed the Informed-RRT* \cite{gammell2014informed} and BIT* (Batch Informed Trees) \cite{gammell2015batch}. Informed-RRT* defines an ellipsoidal region using RRT*'s initial path solution to adaptively sample the configuration space for optimal path planning. Despite improvements in computation time, Informed-RRT* suffers in situations where finding an initial path is itself challenging. On the other hand, BIT* is an incremental graph search method that instantiates a dynamically-changing ellipsoidal region for batch sampling to compute paths. Despite some improvements in computation speed, these biased sampling heuristics still suffer from the curse of dimensionality.



Lazy edge evaluation methods, on the other hand, have shown to exhibit significant improvements in computation speeds by evaluating edges only along the potential path solutions. However, these methods are critically dependent on the underlying edge selector and tend to exhibit limited performance in cluttered environments \cite{houdeep}. Bidirectional path generation improves the algorithm performance in narrow passages but still inherits the limitations of baseline SMPs \cite{qureshi2015intelligent}\cite{tahir2018potentially}.

Reinforcement learning (RL) \cite{sutton1998introduction} has also emerged as a prominent tool to solve continuous control and planning problems \cite{lillicrap2015continuous}. RL considers Markov Decision Processes (MDPs) where an agent interacts with the environment by observing a state and taking an action which leads to a next state and reward. The reward signal encapsulates the underlying task and provides feedback to the agent on how well it is performing. Therefore, the objective of an agent is to learn a policy to maximize the overall expected reward function. In the past, RL was only employed for simple problems with low-dimensions \cite{deisenroth2013survey} \cite{kober2013reinforcement}. Recent advancements have led to solving harder, higher dimensional problems by merging the traditional RL with expressive function approximators such neural networks, now known as Deep RL (DRL) \cite{mnih2015human} \cite{duan2016benchmarking}. DRL has solved various challenging robotic tasks using both model-based \cite{levine2016end} \cite{levine2018learning} \cite{yahya2017collective} and model-free \cite{chebotar2017path} \cite{gu2017deep} \cite{popov2017data} approaches. Despite considerable progress, solving practical problems which have weak rewards and long-horizons remain challenging \cite{qureshi2020composing}. 

There also exist approaches that apply various learning strategies such as imitation learning to mitigate the limitations of motion planning methods. Zucker et al. \cite{zucker2008adaptive} proposed to adapt the sampling for the SMPs using REINFORCE algorithm \cite{williams1992simple} in discretized workspaces. Berenson et al. \cite{berenson2012robot} use a learned heuristic to store new paths, if needed, or to recall and repair the existing paths. Coleman et al. \cite{coleman2015experience} store experiences in a graph rather than individual trajectories. 
Ye and Alterovitz \cite{ye2017guided} use human demonstrations in conjunction with SMPs for path planning. While improved performance was noted compared to traditional planning algorithms, these methods lack generalizability and require tedious hand-engineering for every new environment. Therefore, modern techniques use efficient function approximators such as neural networks to either embed a motion planner or to learning auxiliary functions for SMPs such as sampling heuristic to speed up planning in complex cluttered environments.

Neural Motion Planning has been a recent addition to the motion planning literature. Bency et al. \cite{bency2019neural} introduced recurrent neural networks to embed a planner based on its demonstrations. While useful for learning to navigate static environments, their method does not use environment information and therefore, is not meant to generalize to other environments. Ichter et al. \cite{ichter2018learning} proposed conditional variational autoencoders that contextualize on environment information to generate samples through decoder network for the SMPs such as FMT* \cite{janson2016fast}. The SMPs use the generated samples to create a graph for finding a feasible or optimal path for the robot to follow. In a similar vein, Zhang et. al \cite{zhang2018learning} learns a rejection sampling policy that rejects or accepts the given uniform samples before making them the part of the SMP graph. The rejection sampling policy is learned using past experiences from the similar environments. Note that a policy for rejection sampling implicitly learns the sampling distribution whereas Icheter et. al \cite{ichter2018learning} explicitly generates the guided samples for the SMPs. Bhardwaj et al \cite{bhardwaj2017learning} proposed a method called SAIL that learns a deterministic policy which guides the graph expansion of underlying graph-based planner towards the promising areas that potentially contains the path solution. SAIL learns the guiding policy using the oracle Q-values (encapsulates the cost-to-go function), argmin regression, and full environment information. Like these adaptive sampling methods, MPNet can also generate informed samples for SMPs, but in addition, our approach is also outputs feasible trajectories with worst-case theoretical guarantees.

There has also been attempts towards building learning-based motion planners. For instance, Value Iteration Networks (VIN) \cite{tamar2016value} approximates a planner by emulating value iteration using recurrent convolutional neural networks and max-pooling. However, VIN is only applicable for discrete planning tasks. Universal Planning Networks (UPN) \cite{srinivas2018universal} extends VIN to continuous control problems using gradient descent over generated actions to find a motion plan connecting the given start and goal observations. However, these methods do not generalize to novel environments or tasks and thus require frequent retraining. The most relevant approach to our neural planner (MPNet) is L2RRT \cite{ichter2019robot} that plans motion in learned latent spaces using RRT method. L2RRT learns state-space encoding model, agent's dynamics model, and collision checking model. However, it is unclear that existence of a path solution in configuration space will always imply the existence of a path in the learned latent space and vice versa. 

\begin{figure*}
    \centering
   \begin{subfigure}[b]{0.75\textwidth}
       \includegraphics[height=5.5cm]{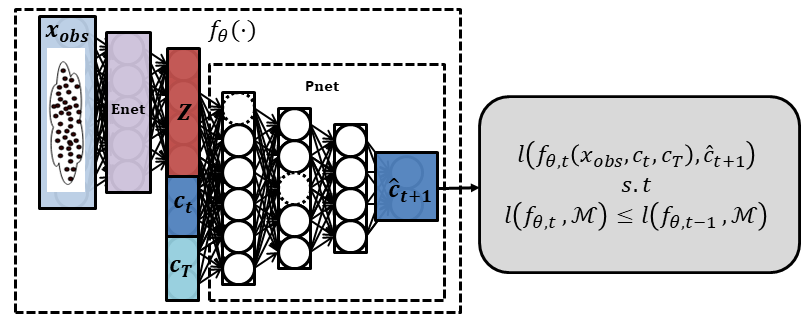}
       \caption{}
    \end{subfigure}
     \begin{subfigure}[b]{0.23\textwidth}
        \includegraphics[height=5.4cm]{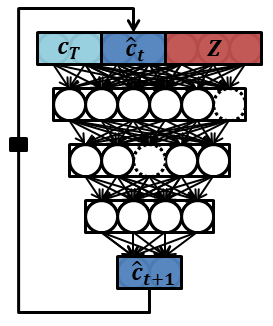}
        \caption{}
    \end{subfigure}
    \caption{MPNet consists of encoder network (Enet) and planning network (Pnet). Fig (a) shows that Pnet and Enet can be trained end-to-end and can learn under continual learning settings from streaming data using constraint optimization and episodic memory $\mathcal{M}$ for a given loss function $l(\cdot)$. Fig (b) shows the online execution of MPNet's neural models. Enet can be a fully-connected neural network or a 3D convolutional neural network that takes the environment information as a point cloud or a 3D voxel $x_\mathrm{obs}$, respectively, and embed them into latent space $Z$. Pnet takes the encoding $Z$, robot current $c_t$ and goal $c_T$ states, and incrementally produces a collision-free trajectory connecting the given start and goal pair.}\label{mpnet2} \vspace*{-4mm}
\end{figure*}

\section{MPNet: A Neural Motion Planner}
MPNet comprises of two neural networks: an encoder network (Enet) and a planning network (Pnet). Enet takes the robot's surrounding information such as a raw point-cloud or point-cloud converted to voxel depending on the underlying neural architecture that could be a fully-connected neural network or a 3D convolutional neural network (CNN), respectively. The output of the Enet is a latent space embedding of the given information. Pnet takes the encoding of the environment, the robot's current state and goal state to output samples for either a path or tree generation. In remaining section, we describe the notations necessary to outline MPNet.

Let robot configuration space (C-space) be denoted as $\mathcal{C} \subset \mathbb{R}^d$ comprising of obstacle space $\mathcal{C}_\mathrm{obs}$ and obstacle-free space $\mathcal{C}_\mathrm{free}=\mathcal{C} \backslash \mathcal{C}_\mathrm{obs}$, where $d$ is the C-space dimensionality. Let robot's surrounding environment, also known as workspace, be denoted as $\mathcal{X} \subset \mathbb{R}^m$, where $m$ is a workspace dimension. Like C-space, the workspace also comprise of obstacle, $\mathcal{X}_\mathrm{obs}$, and obstacle-free, $\mathcal{X}_\mathrm{free}=\mathcal{X} \backslash \mathcal{X}_\mathrm{obs}$, regions. The workspaces could be up to 3-dimensions whereas the C-space can have higher dimensions depending on the robot's degree-of-freedom (DOF). Let robot initial and goal configuration space be  $c_\mathrm{init} \in \mathcal{C}_\mathrm{free}$ and $c_\mathrm{goal} \subset \mathcal{C}_\mathrm{free}$, respectively. Let $\sigma=\{c_0, \cdots, c_T\}$ be an ordered list of length $T$. We assume $\sigma_i$ corresponds to the $i$-th state in $\sigma$, where $i=[0,T]$. For instance $\sigma_0$ corresponds to state $c_0$. Furthermore, we consider $\sigma_\mathrm{end}$ corresponds to the last element of $\sigma$, i.e., $\sigma_\mathrm{end}=c_T$. A motion planning problem can be concisely denoted as $\{c_\mathrm{init}, c_\mathrm{goal}, \mathcal{C}_\mathrm{obs}\}$ where the aim of a planning algorithm is to find a feasible path solution, if one exists, that connects $c_\mathrm{init}$ to $c_\mathrm{goal}$ while completely avoiding obstacles in $\mathcal{C}_\mathrm{obs}$. Therefore, a feasible path can be represented as an ordered list $\sigma=\{c_0, \cdots, c_T\}$ such that $c_0=c_\mathrm{init}$, $c_T=c_\mathrm{goal}$, and a path constituted by connecting consecutive states in $\sigma$ lies entirely in $\mathcal{C}_\mathrm{free}$. In practice, C-space representation of obstacles $\mathcal{C}_\mathrm{obs}$ is unknown and rather a collision-checker is available that takes workspace information, $\mathcal{X}_\mathrm{obs}$,  and robot configuration, $c$, and determines if they are in collision or not. Another important problem in motion planning is to find an optimal path solution for a given cost function. Let a cost function be defined as $J(\cdot)$. An optimal planner provides guarantees, either weak or strong, that if given enough running time, it would eventually converge to an optimal path, if one exists. The optimal path is a solution that has the lowest possible cost w.r.t. $J(\cdot)$.

We consider a practical scenario, where MPNet plans feasible, near-optimal paths using raw point-cloud/voxel data of obstacles $x_\mathrm{obs} \subset \mathcal{X}_\mathrm{obs}$. However, like other planning algorithms, we do assume an availability of a collision-checker that verifies the feasibility of MPNet generated paths based on $\mathcal{X}_\mathrm{obs}$. Precisely, Enet, with parameters $\theta^e$, takes the environment information $x_\mathrm{obs}$ and compresses them into a latent space $Z$,
\begin{equation}
Z \gets \mathrm{Enet}(x_\mathrm{obs}; \theta^e)
\end{equation}
Pnet, with parameters $\theta^p$, takes the environment encoding $Z$, robot's current or initial configuration $c_t \in \mathcal{C}_\mathrm{free}$, and goal configuration $c_\mathrm{goal} \subset \mathcal{C}_\mathrm{free}$ to produce a trajectory through incremental generation of states $\hat{c}_{t+1}$ (Fig. \ref{mpnet2} (b)),
\begin{equation}
\hat{c}_{t+1} \gets \mathrm{Pnet} (Z, c_{t}, c_\mathrm{goal}; \theta^p)
\end{equation} 
\vspace*{-0.5in} 
\\ \\ 
\section{MPNet: Training}
In this section, we present three training methodologies for MPNet neural models, Enet and Pnet: i) offline batch learning, ii) continual learning, and iii) active continual learning. 


Offline batch learning method assumes the availability of complete data to train MPNet offline before running it online to plan motions for unknown/new planning problems. Continual learning enables MPNet to learn from streaming data without forgetting past experiences. Active continual learning incorporates MPNet into the continual learning process where MPNet actively asks for an expert demonstration when needed for the given problem. Further details on training approaches are presented as follow.

\subsection{Offline batch learning}
The offline batch learning requires all the training data to be available in advance for MPNet \cite{qureshi2018motion} \cite{qureshi2018deeply}. As mentioned earlier, MPNet comprises of two modules, Enet and Pnet. In this training approach, both modules can either be trained together in an end-to-end fashion using planning network loss function or separately with their individual loss functions.

To train Enet and Pnet together, we back-propagate the gradient of Pnet's loss function in Equation \ref{mse} through both modules. For the standalone training of Enet, we use an encoder-decoder architecture whenever there is an ample amount of environmental point-cloud data available for unsupervised learning. There exist several techniques for encoder-decoder training such as variational auto-encoders \cite{kingma2013auto} and their variants \cite{higgins2017beta} or the class of contractive autoencoders (CAE) \cite{rifai2011contractive}. For our purpose, we observed that the contractive autoencoders learn robust feature spaces desired for planning and control, and give better performance than other available encoding techniques. The CAE uses the usual reconstruction loss and a regularization over encoder parameters,  
\begin{equation}\vspace*{-0.1in} \label{obl}
l_\mathrm{AE}\big(\theta^e,\theta^d\big)= \cfrac{1}{N_\mathrm{obs}}\sum_{x\in D_\mathrm{obs}}||x-\hat{x}||^2  + \lambda \sum_{ij} (\theta^e_{ij})^2
\end{equation}
where  $\theta^e$, $\theta^d$ are the encoder and decoder parameters, respectively, and $\lambda$ denotes regularization coefficient. The variable $\hat{x}$ represents reconstructed point-cloud. The training dataset of obstacles' point-cloud $x \subset \mathcal{X}_\mathrm{obs}$ is denoted as $D_\mathrm{obs}$ which contains point cloud from $N_\mathrm{obs} \in \mathbb{N}$ different workspaces. 

The planning module (Pnet) is a stochastic feed-forward deep neural network with parameters $\theta^p$. 
Our demonstration trajectory is a list of waypoints, $\sigma=\{c_0,c_1,\cdots,c_T\}$, connecting the start and goal configurations such that the fully connected path lies in obstacle-free space. To train Pnet either end-to-end with Enet or separately for the given expert trajectory, we consider one-step look ahead prediction strategy. Therefore, MPNet takes the obstacles' point-cloud embedding $Z$, robot's current state $c_t$ and goal state $c_T$ as an input to output the next waypoint $\hat{c}_{t+1}$ towards the goal-region. The training objective is a mean-squared-error (MSE) loss between the predicted $\hat{c}_{t+1}$ and target waypoints $c_{t+1}$, i.e.,
\begin{equation}\vspace*{-0.1in}\label{mse}
l_\mathrm{Pnet}(\boldsymbol{\theta})=\cfrac{1}{N_p} \sum^{\hat{N}}_j \sum^{T_j-1}_{i=0} ||\hat{c}_{j,i+1}-c_{j,i+1}||^2,
\end{equation}
where $T_j$ is the length of $j$-th trajectory, $\hat{N} \in \mathbb{N}$ is the total number of training trajectories, and $N_p$ is the averaging term. Although we use MSE loss, one can consider other choices such as adversarial loss function \cite{goodfellow2014generative}. In rigid body planning, such as planning in SE(3), if the rotations are represented as quaternions, we use the following loss:

\begin{equation}\vspace*{-0.1in}
l_\mathrm{Pnet}(\boldsymbol{\theta})=l_p+\beta l_q,
\end{equation}
where $l_p$ and $l_q$ correspond to the MSE loss between positional $p$ and quaternion $q$ components of the waypoints in $\sigma$, and $\beta$ is a scaling factor. Since quaternion needs to be in a unit sphere the loss $l_q$ is defined as (for more details refer to \cite{kendall2017geometric}): 
\begin{equation}\vspace*{-0.1in}
l_\mathrm{q}=\cfrac{1}{N_p} \sum^{\hat{N}}_j \sum^{T-1}_{i=0} \left\Vert\cfrac{\hat{q}_{j,i+1}}{||\hat{q}_{j,i+1}||}-q_{j,i+1}\right\Vert^2
\end{equation}

\subsection{Continual Learning}
In continual learning settings, both modules of MPNet (Enet and Pnet) are trained in an end-to-end fashion, since both neural models need to adapt to the incoming data (Fig. \ref{mpnet2}(a)). We consider a supervised learning problem. Therefore, the data comes with targets, i.e.,
\begin{equation*}
(s_1,y_1,\cdots,s_i,y_i, \cdots, s_N,y_N)
\end{equation*}
where $s=(c_t,c_T,x_\mathrm{obs})$ is the input to MPNet comprising of the robot's current state $c_t$, the goal state $c_T$, and obstacles information $x_\mathrm{obs}$. The target $y$ is the next state $c_{t+1}$ in the expert trajectory given by an oracle planner. Generally, continual learning using neural networks suffers from the issue of catastrophic forgetting since taking a gradient step on a new datum could erase the previous learning. The problem of catastrophic forgetting is also described in terms of knowledge transfer in the backward direction. 

To describe backward transfer, we introduce few new notations as follow. Let a mean model success rate on a dataset $\mathcal{M}$ after learning from an expert example at time $t-1$ and $t$ be denoted as $\mathcal{A}_{\mathcal{M},t-1}$ and $\mathcal{A}_{\mathcal{M},t}$, respectively.  A metric described as a backward transfer can be written as $\mathcal{B}=\mathrm{A}_{\mathcal{M},t}-\mathrm{A}_{\mathcal{M},t-1}$. A positive backward transfer $\mathcal{B} \geq 0$ indicates that after learning from a new experience, the learning-based planner performed better on the previous tasks represented as $\mathcal{M}$. A negative backward transfer $\mathcal{B}<0$ indicates that on learning from new experience the model  on previous tasks deteriorated.


To overcome negative backward transfer leading to catastrophic forgetting, we employ the Gradient Episodic Memory (GEM) method for lifelong learning \cite{lopez2017gradient}. GEM uses the episodic memory $\mathcal{M}$ that has a finite set of continuum data seen in the past to ensure that the model update doesn't lead to negative backward transfer while allowing only the positive backward transfer. For MPNet, we adapt GEM for the regression problem using the following optimization objective function.    

\begin{multline}\label{gem1}
\min_\theta l(f^t_\theta(s),y) \text{ s.t }\\
\hat{\mathbb{E}}_{(s,y)\sim \mathcal{M}}[l(f^t_\theta(s),y)] \leq \hat{\mathbb{E}}_{(s,y)\sim \mathcal{M}}[l(f^{t-1}_\theta(s),y)]  
\end{multline}
where $l=\| f^t_\theta(s) - y \|^2$ is a squared-error loss,  $f^{t}_\theta$ is the MPNet model at time step $t$ (see Fig. \ref{mpnet2}). Furthermore, note that, if the angle between proposed gradient $(g)$ at time $t$, and the gradient over $\mathcal{M}$ $(g_\mathcal{M})$ is positive, i.e., $\langle g, g_\mathcal{M}\rangle \geq 0$, there is no need to maintain the old function parameters $f^{t-1}_\theta$ because the above equation can be formulated as:
\begin{equation}\label{gem2}
\langle g, g_\mathcal{M} \rangle := \bigg \langle \bigtriangledown_\theta l \big(f_\theta(s),y \big), \hat{\mathbb{E}}_{(s,y)\sim \mathcal{M}}\bigtriangledown_\theta l \big(f_\theta(s),y \big) \bigg \rangle
\end{equation}
where $\hat{\mathbb{E}}$ denotes empirical mean. 

In most cases, the proposed gradient $g$ violates the constraint $\langle g, g_\mathcal{M}\rangle \geq 0$, i.e., the proposed gradient update $g$ will cause increase in the loss over previous data. To avoid such violations, David and Razanto \cite{lopez2017gradient} proposed to project the gradient $g$ to the nearest gradient $g'$ that keeps $\langle g',g_\mathcal{M} \rangle \geq 0$, i.e,  
\begin{equation}\label{gem3}
\min_{g'} \cfrac{1}{2} \| g-g' \|^2_2 \text{ s.t } \langle g', g_\mathcal{M} \rangle \geq 0
\end{equation}

The projection of proposed gradient $g$ to $g'$ can be solved efficiently using Quadratic Programming (QP) based on the duality principle, for details refer to \cite{lopez2017gradient}. 
\begin{algorithm}[t]
\DontPrintSemicolon 
$\text{Initialize memories: episodic } \mathcal{M}\text{ and replay }\mathcal{B}^*$\;
$\text{Set the replay period } r$\;
$\text{Set the replay batch size } N_B$\;
$\text{Initialize MPNet } f_\theta \text{ with parameters } \theta. $\;

\For{$t=0$ \textbf{to} $T$} {
	
$\{\mathcal{X}_\mathrm{obs}, c_\mathrm{init}, c_\mathrm{goal}\}_t \gets \mathrm{GetPlanningProblem()}$\;
$\sigma \gets \mathrm{GetExpertDemo}(\mathcal{X}_\mathrm{obs}, c_\mathrm{init}, c_\mathrm{goal})$\;
   	  $\mathcal{M} \gets \mathrm{UpdateEpisodicMemory}(\sigma,\mathcal{M})$\;
   $\mathcal{B}^* \gets \mathcal{B}^* \cup \sigma$\;
$g \leftarrow \hat{\mathbb{E}}_{(s,y)\sim \sigma}\bigtriangledown_\theta l \big(f_\theta(s),y \big) $\;
$g_\mathcal{M} \leftarrow \hat{\mathbb{E}}_{(s,y)\sim \mathcal{M}}\bigtriangledown_\theta l \big(f_\theta(s),y \big) $\;

Project $g$ to $g'$ using QP based on Equation \ref{gem3}\;
Update parameters $\theta$ w.r.t. $g'$\;   
  \If{$\mathcal{B}^*.\mathrm{size}()>N_B \textbf{ and } \textbf{not } t\textbf{ mod }r$} 
   {
   
   	  $\mathcal{B} \gets \mathrm{SampleReplayBatch(\mathcal{B}^*)}$\;
   	  $g \leftarrow \hat{\mathbb{E}}_{(s,y)\sim \mathcal{B}}\bigtriangledown_\theta l \big(f_\theta(s),y \big) $\;
$g_\mathcal{M} \leftarrow \hat{\mathbb{E}}_{(s,y)\sim \mathcal{M}}\bigtriangledown_\theta l \big(f_\theta(s),y \big) $\;

Project $g$ to $g'$ using QP based on Equation \ref{gem3}\;
Update parameters $\theta$ w.r.t. $g'$\;

   }
   }
\caption{Continual Learning}
\label{algo:CL}
\end{algorithm}

Various data parsing methods are available to update the episodic memory $\mathcal{M}$. These sample selection strategies for episodic memory play a vital role in the performance of continual/life-long learning methods such as GEM \cite{isele2018selective}. There exist several selection metrics such as surprise, reward, coverage maximization, and global distribution matching \cite{isele2018selective}. In our continual learning framework, we use a global distribution matching method to select samples for the episodic memory. For details and comparison of different sample selection approaches, please refer to Section VII-B. The global distribution matching method, also known as reservoir sampling, uses random sampling techniques to populate the episodic memory. The aim is to approximately capture the global distribution of the training dataset since it is unknown in advance. There are several ways to implement reservoir sampling. The simplest approach accepts the new sample at $i$-th step with probability $\cfrac{|\mathcal{M}|}{i}$ to replace a randomly selected old sample from the memory. In other words, it rejects the new sample at $i$-the step with probability $1-\cfrac{|\mathcal{M}|}{i}$ to keep the old samples in the memory. 

In addition to episodic memory $\mathcal{M}$, we also maintain a replay/rehearsal memory $\mathcal{B}^*$. The replay buffer lets MPNet rehearse by learning again on the old samples (Line 14-19). We found that rehearsals further mitigate the problem of catastrophic forgetting and leads to better performance, as also reported by Rolnick et al. \cite{rolnick2018experience} in reinforcement learning setting. Note that replay or rehearsal on the past data is done with the interval of replay period $r \in \mathbb{N}_{\geq 0}$. 

Finally, Algorithm \ref{algo:CL} presents the continual learning framework where an expert provides a single demonstration at each time step, and MPNet model parameters are updated according to the projected gradient $g'$.
\begin{algorithm}[t]
\DontPrintSemicolon 
$\text{Initialize memories: episodic } \mathcal{M} \text{ and replay }\mathcal{B}^*$\;
$\text{Initialize MPNet } f_\theta \text{ with parameters } \theta. $\;
$\text{Set the number of iterations } C \text{ to pretrain MPNet.}$\;
$\text{Set the replay period } r$\;
$\text{Set the replay batch size } N_B$\;

\For{$t=0$ \textbf{to} $T$} {
	
$\{\mathcal{X}_\mathrm{obs}, c_\mathrm{init}, c_\mathrm{goal}\}_t \gets \mathrm{GetPlanningProblem()}$\;
$\sigma \gets \varphi$ $\backslash \backslash$ an empty list to store path solution\;
   \If{$t>N_c$}
   {
   	  $x_\mathrm{obs} \subset \mathcal{X}_\mathrm{obs}$\;
   	  $\sigma \gets f_\theta(x_\mathrm{obs}, c_\mathrm{init}, c_\mathrm{goal})$\;
   }
   \If{$\textbf{not }\sigma$} 
   {
   
   	  $\sigma \gets \mathrm{GetExpertDemo}(\mathcal{X}_\mathrm{obs}, c_\mathrm{init}, c_\mathrm{goal})$\;
   $\mathcal{B}^* \gets \mathcal{B}^* \cup \sigma$\;
   	  $\mathcal{M} \gets \mathrm{UpdateEpisodicMemory}(\sigma,\mathcal{M})$\;
$g \leftarrow \hat{\mathbb{E}}_{(s,y)\sim \sigma}\bigtriangledown_\theta l \big(f_\theta(s),y \big) $\;
$g_\mathcal{M} \leftarrow \hat{\mathbb{E}}_{(s,y)\sim \mathcal{M}}\bigtriangledown_\theta l \big(f_\theta(s),y \big) $\;

Project $g$ to $g'$ using QP based on Equation \ref{gem3}\;
Update parameters $\theta$ w.r.t. $g'$\;   	  
   	  
   }
  \If{$\mathcal{B}^*.\mathrm{size}()>N_B \textbf{ and } \textbf{not } t\textbf{ mod }r$} 
   {
   
   	  $\mathcal{B} \gets \mathrm{SampleReplayBatch(\mathcal{B}^*)}$\;
   	  $g \leftarrow \hat{\mathbb{E}}_{(s,y)\sim \mathcal{B}}\bigtriangledown_\theta l \big(f_\theta(s),y \big) $\;
$g_\mathcal{M} \leftarrow \hat{\mathbb{E}}_{(s,y)\sim \mathcal{M}}\bigtriangledown_\theta l \big(f_\theta(s),y \big) $\;

Project $g$ to $g'$ using QP based on Equation \ref{gem3}\;
Update parameters $\theta$ w.r.t. $g'$\;
   	  
   }
   }
\caption{Active Continual Learning}
\label{algo:ACL}
\end{algorithm}
\subsection{Active Continual Learning}
Active continual learning (ACL) is our novel data-efficient learning strategy which is practical in the sense that the planning problem comes in streams and our method asks for expert demonstrations only when needed. It builds on the framework of continual learning presented in the previous section. ACL introduces a two-level of sample selection strategy. First, ACL gathers the training data by actively asking for the demonstrations on problems where MPNet failed to find a path. Second, it employs a sample selection strategy that further prunes the expert demonstrations to fill episodic memory so that it approximates the global distribution from streaming data. The two-level of data selection in ACL improves the training samples efficiency while learning the generalized neural models for the MPNet.

Algorithm \ref{algo:ACL} presents the outline of ACL method. At every time step $t$, the environment generates a planning problem ($c_\mathrm{init}, c_\mathrm{goal}, \mathcal{X}_\mathrm{obs})$ comprising of the robot's initial $c_\mathrm{init}$ and goal $c_\mathrm{goal}$ configurations and the obstacles' information $\mathcal{X}_\mathrm{obs}$ (Line 7). Before asking MPNet to plan a motion for a given problem, we let it learn from the expert demonstrations for up to $N_c \in \mathbb{N}_{\geq 0}$ iterations (Line 9-10). If MPNet is not called or failed to determine a solution, an expert-planner is executed to determine a path solution $\sigma$ for a given planning problem (Line 13).  The expert trajectory $\sigma$ is stored into a replay buffer $\mathcal{B}^*$ and an episodic $\mathcal{M}$ memory based on their sample selection strategies (Line 14-15). MPNet is trained (Line 16-19) on the given demonstration using the constraint optimization mentioned in Equations \ref{gem1}-\ref{gem3}. Finally, similar to continual learning, we also perform rehearsals on the old samples (Line 20-25).
\begin{figure*}[t]
    \centering
    \begin{subfigure}[b]{0.19\textwidth}
       \includegraphics[width=3.5cm]{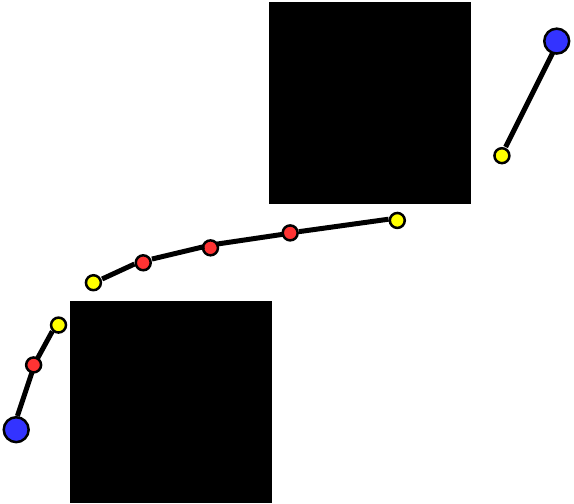}
        \caption{Global Planning}
    \end{subfigure}
    \begin{subfigure}[b]{0.19\textwidth}
       \includegraphics[width=3.5cm]{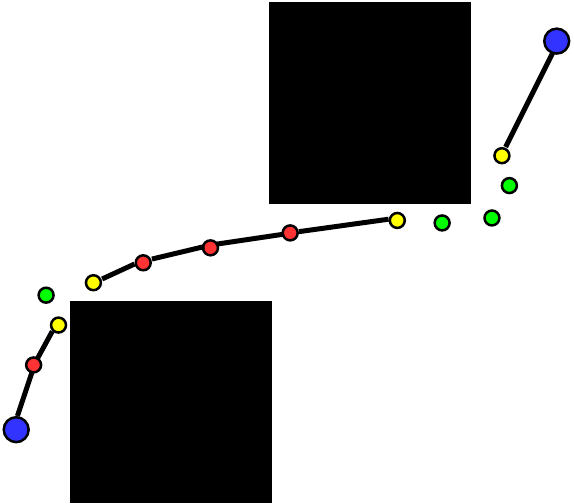}
        \caption{Neural Replanning}
    \end{subfigure}
    \begin{subfigure}[b]{0.19\textwidth}
       \includegraphics[width=3.5cm]{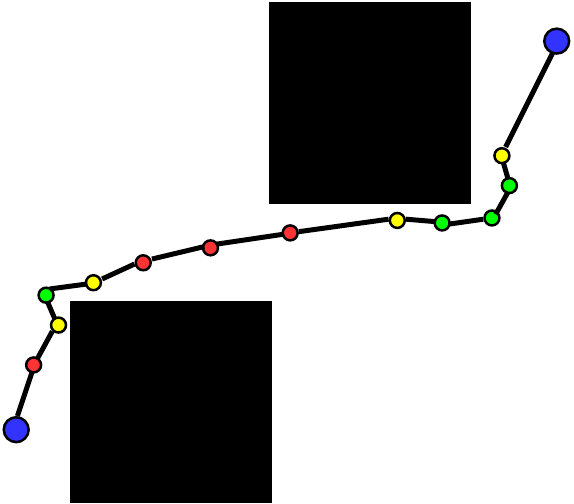}
        \caption{Steering}
    \end{subfigure}
        \begin{subfigure}[b]{0.19\textwidth}
      \includegraphics[width=3.5cm]{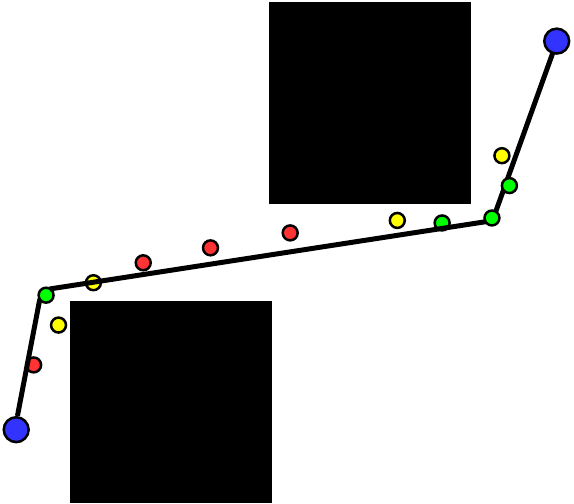}
       \caption{LazyStatesContraction}
    \end{subfigure}
    \begin{subfigure}[b]{0.19\textwidth}
       \includegraphics[width=3.5cm]{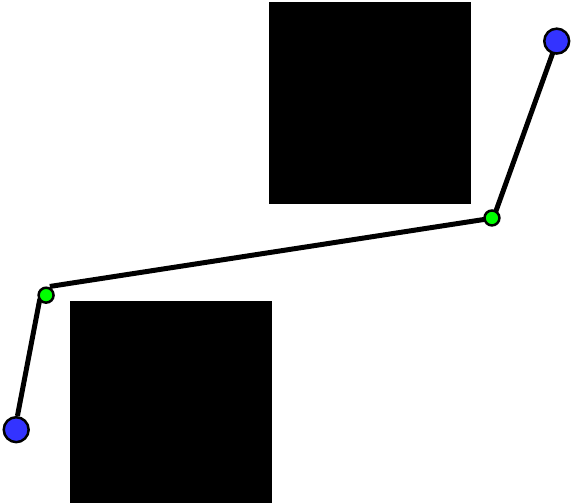}
        \caption{Optimized Path}
    \end{subfigure}
    \caption{Online execution of MPNet to find a path between the given start and goal states (shown in blue dots) in a 2D environment. Our iterative bidirectional planning algorithm uses the trained neural models, Enet and Pnet, to plan the motion. Fig. (a) shows global planning where MPNet outputs a coarse path. The coarse path might contain beacon states (shown as yellow dots) that are not directly connectable as the connecting trajectory passes through the obstacles. Figs. (b-c) depict a replanning step, where the beacon states are considered as the start and goal, and MPNet is executed to plan a motion between them on a finer level, thanks to the Pnet's stochasticity, due to Dropout, that helps in recovery from failures. Fig. (d) shows our lazy states contraction (LSC) method that prunes out the redundant states leading to a lousy path.  Fig. (e) shows the final feasible plan given to the robot to follow by MPNet.}\label{mpnet3}
\end{figure*}
\section{MPNet: Online Path Planning}
MPNet can generate both end-to-end collision-free paths and informed samples for the SMPs. We denote our path planner and sample generator as MPNetPath and MPNetSMP, respectively. MPNet uses two trained modules, Enet and Pnet.
 
Enet takes the obstacles' information $x_\mathrm{obs}$ and encodes them into a latent-space embedding $Z$. Enet is either trained end-to-end with Pnet or separately with encoder-decoder structure. Pnet, is a stochastic model as during execution it uses Dropout in almost every layer. The layers with Dropout \cite{srivastava2014dropout} get their neurons dropped with a probability $p \in [0,1]$. Therefore, every time MPNet calls Pnet,  due to Dropout, it gets a sliced/thinner model of the original planning network which leads to a stochastic behavior. 
The stochasticity helps in recursive, divide-and-conquer based path planning, and also enables MPNet to generate samples for SMPs. The input to Pnet is a concatenated vector of obstacle-space embedding $Z$, robot current configuration $\hat{c}_t$, and goal configuration $c_T$. The output is the robot configuration $\hat{c}_{t+1}$ for time step $t+1$ that would bring the robot closer to the goal configuration. We iteratively execute Pnet (Fig. \ref{mpnet2}(b)), i.e., the new state $\hat{c}_{t+1}$ becomes the current state $\hat{c}_{t}$ in the next time step and therefore, the path grows incrementally.
\subsection{Path Planning with MPNet}
Path planning with MPNet uses Enet and Pnet in conjunction with our iterative, recursive, and bi-directional planning algorithm. Our planning strategy has the following salient features.\\\\
\textbf{\textit{Forward-Backward Bi-directional Planning:}}
Our algorithm is bidirectional as we run our neural models to plan forward, from start to goal, as well as to plan backward, from goal to start, until both paths meet each other. To connect forward and backward paths we use a greedy RRT-Connect \cite{kuffner2000rrt} like heuristic.\\\\
\textbf{\textit{Recursive Divide-and-Conquer Planning:}}
Our planner is recursive and solves the given path planning problem through divide-and-conquer. MPNet begins with global planning (Fig. \ref{mpnet3} (a)) that results in critical, collision-free states that are vital to generating a feasible trajectory. If any of the consecutive critical nodes in the global path are not connectable (Beacon states), MPNet takes them as a new start and goal, and recursively plans a motion between them (Figs. \ref{mpnet3} (b-c)). Hence, MPNet decomposes the given problem into sub-problems and recursively executes itself on those sub-problems to eventually find a path solution.\\

In the remainder of this section, we describe the essential components of MPNetPath algorithm followed by the overall algorithm execution and outline.\\ 

\subsubsection{Bidirectional Neural Planner (BNP)}
In this section, we formally outline our forward-backward, bidirectional neural planning method, outlined in Algorithm \ref{algo:BNP}. BNP takes the environment representation $Z$, the robot's initial $c_\mathrm{init}$ and target $c_\mathrm{goal}$ configurations as an input. The bidirectional path is generated as two paths, from start to goal $(\sigma_a)$ and from goal to start $(\sigma_b)$, incrementally march towards each other. The paths $\sigma_a$ and $\sigma_b$ are initialized with the robot's start configuration $c_\mathrm{init}$ and goal configuration $c_\mathrm{goal}$, respectively (Line 1). We expand paths in an alternating fashion, i.e., if at any iteration $i$, a path  $\sigma_a$ is extended, then in the next iteration, a path $\sigma_b$ will be extended, and this is achieved by swapping the roles of $\sigma_a$ and $\sigma_b$ at the end of every iteration (Line 10). Furthermore, after each expansion step, the planner attempts to connect both paths through a straight-line, if possible. We use $\mathrm{steerTo}$ (described later) to   perform straight-line connection which makes our connection heuristic greedy. Therefore, like RRT-Connect \cite{kuffner2000rrt}, our method makes the best effort to connect both paths after every path expansion. In case both paths $\sigma^a$ and  $\sigma^b$ are connectable, BNP returns a concatenated path $\sigma$ that comprises of states from $\sigma_a$ and $\sigma_b$ (Line 8-9).  
\begin{algorithm}
\DontPrintSemicolon 
$\sigma^\mathrm{a} \gets \{c_\mathrm{init}\}, \sigma^\mathrm{b} \gets \{c_\mathrm{goal}\}$\;

$\sigma\gets \varnothing$\;
\For{$i \gets 0$ \textbf{to} $N$} {
	
	$c_\mathrm{new} \gets \mathrm{Pnet}\big(Z,\sigma^\mathrm{a}_\mathrm{end},\sigma^\mathrm{b}_\mathrm{end}\big)$\;
	$\sigma^\mathrm{a} \gets \sigma^\mathrm{a} \cup \{c_\mathrm{new}\}$\;
	$\mathrm{Connect} \gets \mathrm{steerTo}\big(\sigma^\mathrm{a}_\mathrm{end},\sigma^\mathrm{b}_\mathrm{end}\big)$\;
   \If{$\mathrm{Connect}$}
   {
   	  $\sigma \gets \mathrm{concatenate}(\sigma^\mathrm{a},\sigma^\mathrm{b})$\;
      \Return{$\sigma$}\;
   }
   $\mathrm{SWAP}(\sigma^\mathrm{a},\sigma^\mathrm{b})$\;
   }

\Return{$\varnothing$}\;
\caption{BidirectionalNeuralPlanner($c_\mathrm{init}, c_\mathrm{goal}, Z$)}
\label{algo:BNP}
\end{algorithm}
  
\subsubsection{Replanning}
The bidirectional neural planner outputs a coarse path of critical points $\sigma$ (Fig. \ref{mpnet3} (a)). If all consecutive nodes in $\sigma$ are connectable, i.e., the trajectories connecting them are in obstacle-free space then there is no need to perform any further planning. However, if there are any beacon nodes in $\sigma$, a replanning is performed to connect them (Fig. \ref{mpnet3} (b-c)). The replanning procedure is presented in Algorithm \ref{algo:RNP}. The trajectory $\sigma$ consists of states $\sigma=\{c_0, c_1,\cdots, c_T\}$ given by BNP. The algorithm uses $\mathrm{steerTo}$ function and iterates over every connective state pairs $\sigma_i$ and $\sigma_{i+1}$ in a given path $\sigma$ to check if there exists a collision-free straight trajectory between them. If a collision-free trajectory does not exist between any of the given consecutive states, a new path is determined between them through replanning. To replan, the beacon nodes are presented to the replanner as a new start and goal pair together with the obstacles information. We propose two replanning methods:\\\\ (i) \textit{\textbf{Neural replanning (NP):}} NP takes the beacon states and makes a limited number of attempts ($N_r$) to find a path solution between them using BNP. In case of NP, the $\mathrm{plan\_oracle}$ in Algorithm \ref{algo:RNP} is set to $\mathrm{False}$.\\\\(ii) \textit{\textbf{Hybrid replanning (HP):}} HP combines NP and oracle planner. HP takes beacon states and tries to find a solution using NP. If NP fails after a fixed number of trials $N_r$, an oracle planner is executed to find a solution, if one exists, between the given states (Line 9). HP is performed if boolean $\mathrm{plan\_oracle}$ is set $\mathrm{True}$ in Algorithm \ref{algo:RNP}.\\  
\begin{algorithm}[t]
\DontPrintSemicolon 
$\sigma_\mathrm{new}\gets \varnothing$\;
\For{$i \gets 0$ \textbf{to} $size(\sigma)-1$} {
   		\If{$\mathrm{steerTo}(\sigma_i,\sigma_{i+1})$}
   		{
   			$\sigma_\mathrm{new} \gets \sigma_\mathrm{new} \cup \{\sigma_i,\sigma_{i+1}\}  $\;
   		}
   		\Else
   		{   		
   		\If{$\mathrm{plan\_oracle}$}
   		{
   		$\sigma' \gets \mathrm{OraclePlanner}(\sigma_i,\sigma_{i+1},\mathcal{X}_\mathrm{obs})$\;
   		}
   		\Else
   		{

   		  $\sigma' \gets \mathrm{BNP}(\sigma_i,\sigma_{i+1},Z)$\;
   		}
   		
   		\If{$\sigma'$}{
   			$\sigma_\mathrm{new} \gets \sigma_\mathrm{new} \cup \sigma'$\;
   		}
   		\Else
   		{
   		
   		\Return$\varnothing$\; 
   		}
   		}

   	}
   	\Return$\sigma_\mathrm{new}$\;
\caption{Replan($\sigma, Z, \mathrm{plan\_oracle}$)}
\label{algo:RNP}
\end{algorithm} 
 
\subsubsection{Lazy States Contraction (LSC)}
This process is often known as smoothing or shortcutting \cite{hauser2010fast}. The term contraction was coined in graph theory literature \cite{skienaimplementing}. We implement LSC as a recursive function that when executed on a given path $\sigma=\{c_0, c_1, \cdots, c_T\}$, leaves no redundant state by directly connecting, if possible, the non-consecutive states, i.e., $c_i$ and $c_{>i+1}$, where $i \in [0, T-1]$, and removing the lousy/lazy states (Fig. \ref{mpnet3} (d)). This method improves the computational efficiency as the algorithm will have to process fewer nodes during planning.
\subsubsection{Steering ($\mathrm{SteerTo}$)}
The $\mathrm{steerTo}$ function, as it name suggests, walks through a straight line connecting the two given states to validate if their connecting trajectory lies entirely in the collision-free space or not. To evaluate the connecting trajectory, the $\mathrm{steerTo}$ function discretize the straight line into small steps and verifies if each discrete node is collision-free or not. The discretization is done as $\sigma(\delta) = (1 - \delta) c_1 + \delta c_2; \forall\delta \in [0, 1]$ between nodes $c_1$ and $c_2$ using a small step-size. In our algorithm, we use different step-sizes for the global planning, replanning, and for final feasibility checks (for details refer to supplementary material).

\subsubsection{isFeasible}
This function uses the $\mathrm{steerTo}$ to check whether the given path $\sigma=\{c_0, c_1, \cdots, c_T\}$ lies entirely in collision-free space or not.\\

\begin{algorithm}
\DontPrintSemicolon 
$Z \gets \mathrm{Enet}(x_\mathrm{obs})$\;
$\sigma \gets \mathrm{BNP}(c_\mathrm{init}, c_\mathrm{goal}, Z)//$ \small{BidirectionalNeuralPlanner}\;

$\sigma \gets \mathrm{LazyStatesContraction}(\sigma)$\;
 \If{$\mathrm{IsFeasible}(\sigma)$}
   {
      \Return$\sigma$\
   }
   \Else
   {
   {$\mathrm{plan\_oracle}=\mathrm{False}//$} \small{use NeuralReplanning} \;
   \For{$i \gets 0$ \textbf{to} $N_r$}{
   	$\sigma \gets \mathrm{Replan}(\sigma,Z,\mathrm{plan\_oracle})$\;
   	$\sigma \gets \mathrm{LazyStatesContraction}(\sigma)$\;
   	
 \If{$\mathrm{IsFeasible}(\sigma)$}
   {
      \Return$\sigma$\;
   }
   }
   {$\mathrm{plan\_oracle}=\mathrm{True}//$} \small{use OraclePlanner} \;
   $\sigma \gets \mathrm{Replan}(\sigma,Z,\mathrm{plan\_oracle})//$ \small{HybridReplanning}\;
   $\sigma \gets \mathrm{LazyStatesContraction}(\sigma)$\;
   \If{$\mathrm{IsFeasible}(\sigma)$}
   {
      \Return$\sigma$\;
   }
   \Return$\varnothing$\;
   }

\caption{MPNetPath($c_\mathrm{init},c_\mathrm{goal}, x_\mathrm{obs}$)}\label{alg:mpnetpath}
\end{algorithm} 

\textit{\textbf{MPNetPath Execution Summary:}}
Algorithm \ref{alg:mpnetpath} outlines our MPNetPath framework. Enet encodes the raw point-cloud $x_\mathrm{obs}$ into latent embedding $Z$ (Line 1). BNP takes the given planning problem $(c_\mathrm{init}, c_\mathrm{goal}, Z)$ and outputs a coarse path $\sigma$ (Line 2). The LSC function takes $\sigma$ to remove the redundant nodes and leaves only critical states crucial for finding a feasible path solution (Line 3). If a path constituted by connecting the consecutive nodes in $\sigma$ does not belong to the collision-free region, a neural replanning (NP), followed by LSC, is performed for a fixed number of iterations $N_r$ (Line 8-12). The neural-replanning recursively gets deeper and finer in connecting the beacon states whereas LSC function keeps on pruning out the redundant states after every replanning step. In most case, the neural replanner is able to compute a path solution. However, for some hard cases, where neural replanner fails to find a path between beacon states in $\sigma$, we employ hybrid planning (HP) using an oracle planner (Line 13-14). Note that, in case of hybrid replanning, the oracle planner is executed only for a small segment of the overall problem which helps MPNetPath retains its computational benefits while being able to determine a path solution if one exists.

\subsection{Informed Sampling with MPNet}
As mentioned earlier, our planning network (Pnet) is a stochastic model as it has Dropout in almost every layer during the execution. We exploit the stochasticity of Pnet to generate multiple informed samples that would be used as a sample generator for a classical SMP. Because samples are informed in such a way that has high probability to be part of a connectable and near-optimal path, it allows the underlying SMP to find solutions efficiently and quickly. 

Algorithm \ref{algo:mpnetsmp} outlines the procedure to integrate our MPNetSMP with any SMP. MPNet's BNP performs one-step-ahead prediction, i.e., given obstacles representation $Z$, robot current state $\hat{c}_t$, and goal state $c_T$, it predicts a next state $\hat{c}_{t+1}$ closer to the goal than previous state $\hat{c}_t$. We execute MPNetSMP to incrementally generate samples from start to goal in a loop for a fixed number of iterations $N_\mathrm{smp}$ before switching to uniform random sampling (Line 5-8). The underlying SMP takes the informed samples (Line 9) and builds a tree starting from an initial configuration. Due to informed sampling, the tree, in the beginning, is biased towards a subspace that potentially contains a path solution and after certain iterations, the tree begins to expand uniformly. Hence, our MPNetSMP perform both exploitation and exploration. Once a path solution $\sigma$ is found, it is either returned as a solution or further optimized for a given cost function. 

We also propose that our MPNetSMP can be adapted to generate samples for bidirectional SMPs \cite{qureshi2015intelligent}. MPNetSMP generates informed samples incrementally between the given start and goal, and just like the bidirectional heuristic of MPNetPath, it can be queried to generate bidirectional samples. A simple modification for bidirectional sampling in Algorithm \ref{algo:mpnetsmp} would be to initialize two random variable $c_\mathrm{randS}$ and $c_\mathrm{randG}$ with start and goal states, respectively, in place of Line 2. Then use $c_\mathrm{randG}$ instead of $c_\mathrm{goal}$ and swap the roles of $c_\mathrm{randS}$ and $c_\mathrm{randG}$ at the end of every iteration. The proposed modification would allow informed bidirectional generation of trees that are crucial for solving the narrow passage planning problems.  

\begin{algorithm}[t]
\DontPrintSemicolon 
\textbf{Initialize} SMP($c_\mathrm{init},c_\mathrm{goal}, \mathcal{X}_\mathrm{obs}$)\;
$c_\mathrm{rand} \gets c_\mathrm{init}$\;
$Z \gets \mathrm{Enet}(x_\mathrm{obs})$\;
\For{$i \gets 0$ \textbf{to} $n$} {

\If{$i <N_\mathrm{smp}$}
   {
   $c_\mathrm{rand} \gets \mathrm{BNP}\big(Z, c_\mathrm{rand}, c_\mathrm{goal}\big)$\;
   	  
   }
 \Else
 {
    $c_\mathrm{rand} \gets \mathrm{RandomSampler}()$\;

 }

    $\sigma \gets \mathrm{SMP}\big(c_\mathrm{rand}\big)$\;
	
   \If{$c_\mathrm{rand} \in c_\mathrm{goal}$}
   {
   $c_\mathrm{rand} \gets c_\mathrm{init}$\;
   	  
   }
   }

   \If{$\sigma_\mathrm{end} \in c_\mathrm{goal}$}
   {
   \Return{$\sigma$}\;
   	  
   }

\Return{$\varnothing$}\;
\caption{MPNetSMP($c_\mathrm{init}, c_\mathrm{goal}, x_\mathrm{obs} $)}
\label{algo:mpnetsmp}
\end{algorithm}

\begin{figure*}[t]
    \centering
    \begin{subfigure}[b]{0.23\textwidth}
      \includegraphics[height=3.9cm,width=4.1cm]{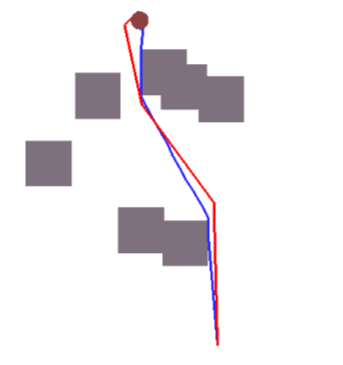}
       \caption{$t_R=4.96s, t_{MP}=0.02s$\\\hspace*{0.17in}$ c_R=38.23, c_{MP}=39.17$}
    \end{subfigure}
    \begin{subfigure}[b]{0.23\textwidth}
       \includegraphics[height=3.9cm,width=4.1cm]{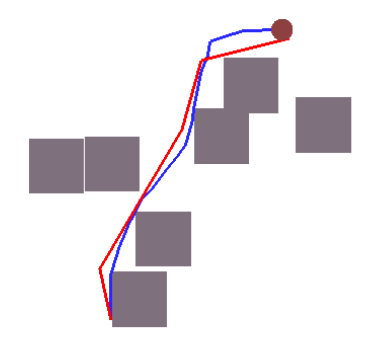}
        \caption{$t_R=5.35s, t_{MP}=0.02s$\\\hspace*{0.17in}$ c_R=34.09, c_{MP}=34.31$}
    \end{subfigure}
    \begin{subfigure}[b]{0.23\textwidth}
       \includegraphics[height=3.85cm,width=4.1cm]{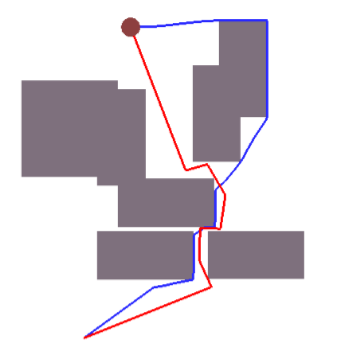}
        \caption{$t_R=6.83s, t_{MP}=0.03s$\\\hspace*{0.17in}$ c_R=45.52, c_{MP}=44.31$}
    \end{subfigure}
    \begin{subfigure}[b]{0.23\textwidth}
       \includegraphics[height=3.85cm,width=4.1cm]{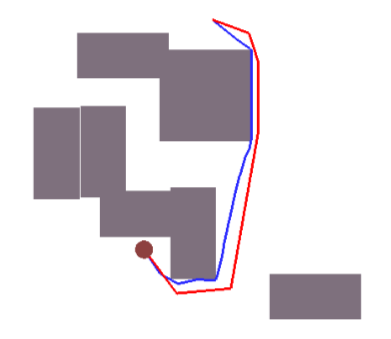}
        \caption{$t_R=6.22s, t_{MP}=0.04s$\\\hspace*{0.17in}$ c_R=40.93, c_{MP}=41.79$}
    \end{subfigure}
        \begin{subfigure}[b]{0.23\textwidth}
      \includegraphics[height=3.9cm,width=4.1cm]{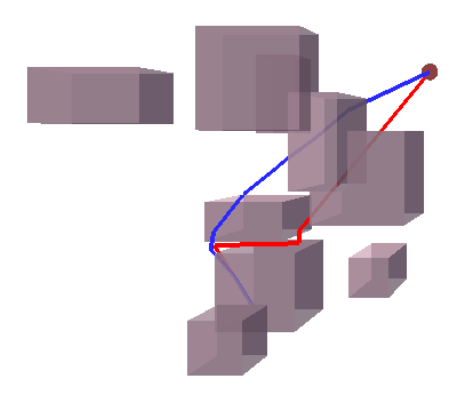}
       \caption{$t_R=8.37s, t_{MP}=0.07s$\\\hspace*{0.17in}$ c_R=49.38, c_{MP}=51.37$}
    \end{subfigure}
    \begin{subfigure}[b]{0.23\textwidth}
       \includegraphics[height=3.9cm,width=4.1cm]{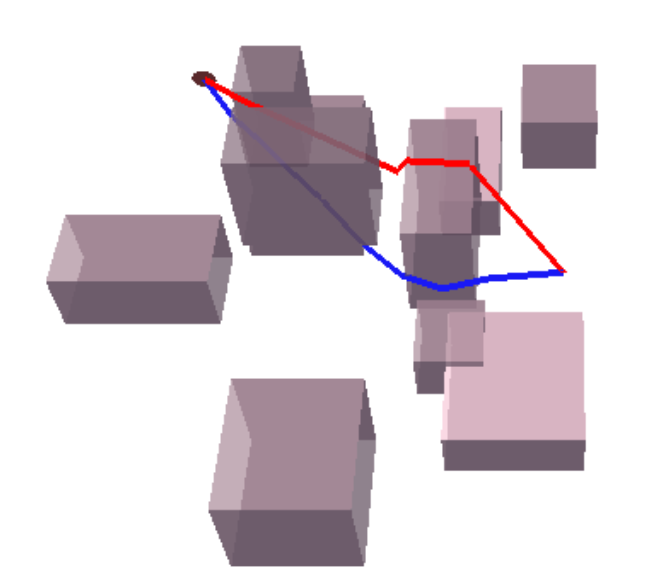}
        \caption{$t_R=9.70s, t_{MP}=0.08s$\\\hspace*{0.17in}$ c_R=43.35, c_{MP}=43.12$}
    \end{subfigure}
    \begin{subfigure}[b]{0.23\textwidth}
       \includegraphics[height=3.85cm,width=4.1cm]{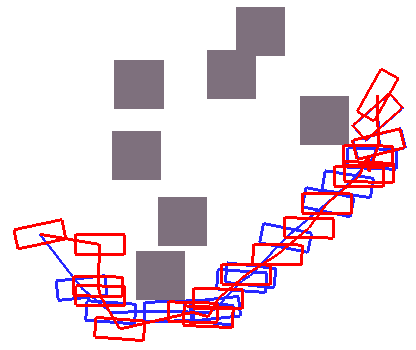}
        \caption{$t_R=26.61s, t_{MP}=0.38s$\\\hspace*{0.18in}$ c_R=42.98, c_{MP}=43.69$}
    \end{subfigure}
    \begin{subfigure}[b]{0.23\textwidth}
       \includegraphics[height=3.85cm,width=4.1cm]{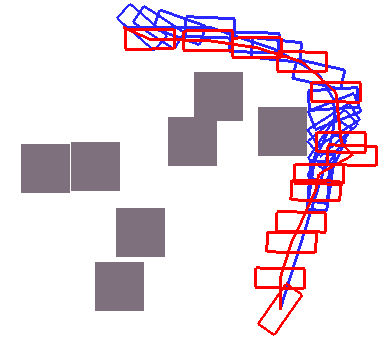}
        \caption{$t_R=27.81s, t_{MP}=0.37s$\\\hspace*{0.18in}$ c_R=39.08, c_{MP}=39.71$}
    \end{subfigure}
    \caption{Time comparison of MPNetPath with neural replanning (Red) and RRT* (Blue) for computing the near-optimal path solutions in example environments. Figs.(a-b) and (c-d) present simple 2D and complex 2D environments. Figs. (e-f) indicates complex 3D case whereas Figs. (g-h) shows the rigid-body-SE2 case, also known as piano mover's problem. In these environments, it can be seen that MPNet plans paths of comparable lengths as its expert demonstrator RRT* but in extremely shorter amounts of time.}\label{mpnetpath}
\end{figure*}

\begin{figure*}[t]
    \centering
    \begin{subfigure}[b]{0.23\textwidth}
      \includegraphics[height=3.9cm,width=3.9cm]{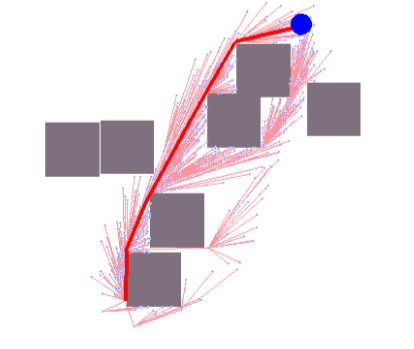}
       \caption{$t=0.13s, c=32.46$}
    \end{subfigure}
    \begin{subfigure}[b]{0.23\textwidth}
       \includegraphics[height=3.9cm,width=3.9cm]{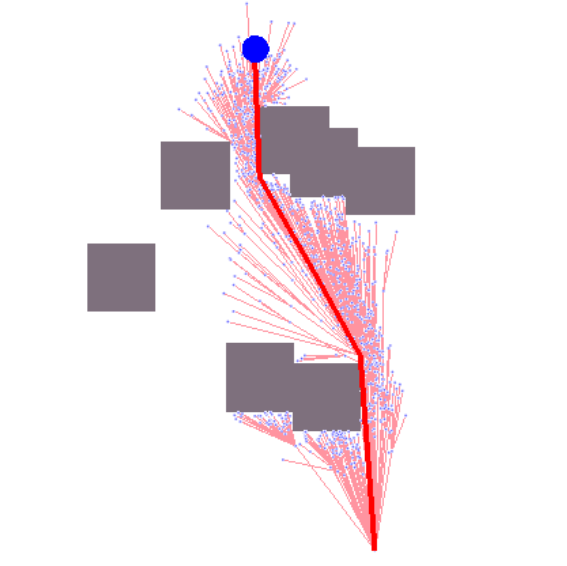}
        \caption{$t=0.13s, c=38.02$}
    \end{subfigure}
    \begin{subfigure}[b]{0.24\textwidth}
       \includegraphics[height=3.85cm,width=4.1cm]{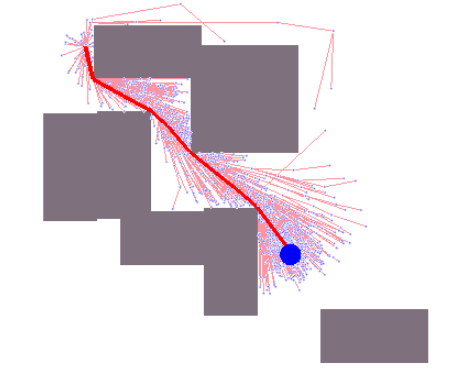}
        \caption{$t=0.25s, c=27.57$}
    \end{subfigure}
    \begin{subfigure}[b]{0.24\textwidth}
       \includegraphics[height=3.85cm,width=4.1cm]{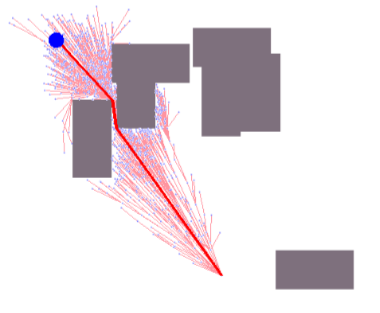}
        \caption{$t=0.21s, c=37.23$}
    \end{subfigure}
        \begin{subfigure}[b]{0.23\textwidth}
      \includegraphics[height=3.9cm,width=3.9cm]{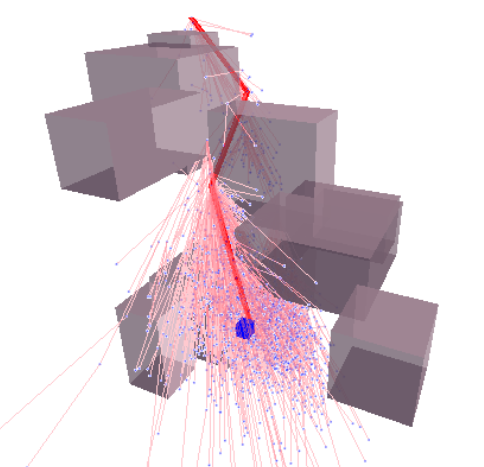}
       \caption{$t=0.22s, c=40.73$}
    \end{subfigure}
    \begin{subfigure}[b]{0.23\textwidth}
       \includegraphics[height=3.9cm,width=3.9cm]{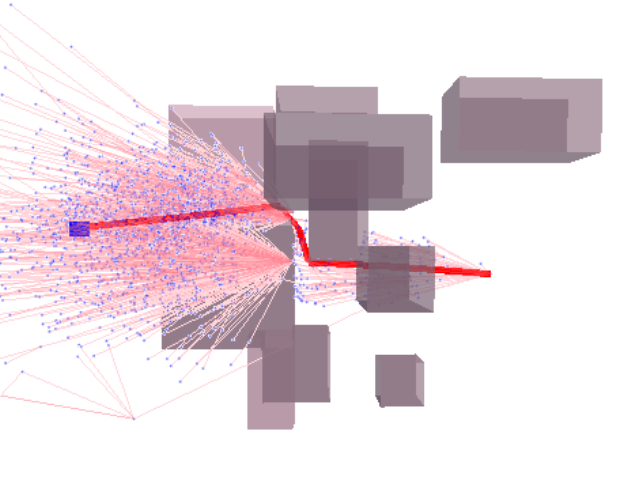}
        \caption{$t=0.18s, c=36.48$}
    \end{subfigure}
    \begin{subfigure}[b]{0.24\textwidth}
       \includegraphics[height=3.85cm,width=4.1cm]{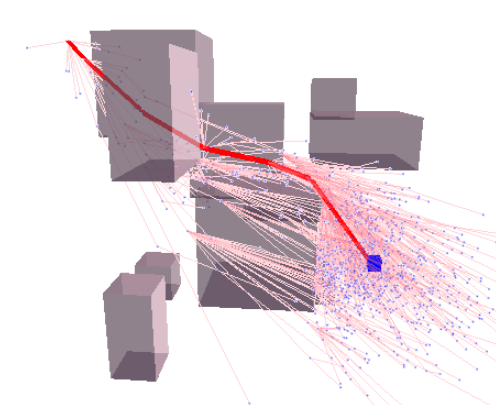}
        \caption{$t=0.29s, c=39.51$}
    \end{subfigure}
    \begin{subfigure}[b]{0.24\textwidth}
       \includegraphics[height=3.85cm,width=4.1cm]{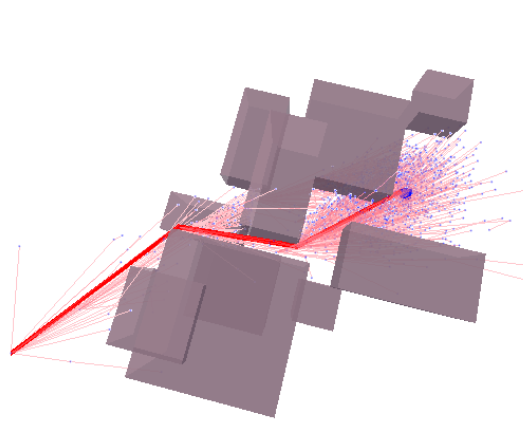}
        \caption{$t=0.21s, c=40.26$}
    \end{subfigure}
    \caption{MPNetSMP generating informed samples for RRT* to plan motions in simple 2D (Figs. a-b), complex 2D (Figs c-d) and complex 3D (Figs. e-h). The number of samples and time required to compute the path are denoted by $n$ and $t$, respectively. Our experiments show that MPNetSMP based RRT* is atleast 30 times faster than traditional RRT* in the presented environments.}\label{mpnetsmp}
\end{figure*}

\begin{table*}[b]
\centering 
\begin{tabular}{|c|c|c|c|c|}\hline
\multirow{2}{*}{Methods}&\multicolumn{4}{c|}{Environments}\\\cline{2-5}
&\multicolumn{1}{c|}{Simple 2D }&\multicolumn{1}{c|}{Complex 2D}&\multicolumn{1}{c|}{Complex 3D}&\multicolumn{1}{c|}{Rigid-body-SE2}\\\hline
\multirow{1}{*}{Informed-RRT*}& \multirow{1}{*}{$1.06\pm 0.33$ $(1.10\pm 0.09)$} & \multirow{1}{*}{$1.60\pm 0.47$ $(1.49\pm 0.16)$}  &\multirow{1}{*}{ $2.99\pm 0.82$ $(2.76\pm 0.20)$}&\multirow{1}{*}{$15.58\pm 2.85$ $(14.80\pm 2.83)$}  \\ \hline

\multirow{1}{*}{BIT*}& \multirow{1}{*}{$0.59\pm 0.28$ $(0.65\pm 0.30)$} & \multirow{1}{*}{$1.79\pm 1.35$ $(1.61\pm 0.53)$}  &\multirow{1}{*}{ $0.19\pm 0.12$ $(0.20\pm 0.04)$}&\multirow{1}{*}{$7.16\pm 1.95$ $(6.52\pm 1.65)$}\\ \hline

\multirow{1}{*}{MPNetSMP (B)}& \multirow{1}{*}{$0.13\pm 0.01$ $(0.13\pm 0.00)$} & \multirow{1}{*}{$0.29\pm 0.05$ $(0.23\pm 0.07)$}  &\multirow{1}{*}{ $0.25\pm 0.05$ $(0.23\pm 0.06)$}&\multirow{1}{*}{$0.49\pm 0.05$ $(0.39\pm 0.04)$} \\ \hline

\multirow{1}{*}{MPNetPath:NP (C)}& \multirow{1}{*}{$0.02\pm 0.00$ $(0.02\pm 0.00)$} & \multirow{1}{*}{$0.05\pm 0.01$ $(0.05\pm 0.01)$}  &\multirow{1}{*}{ $0.07\pm 0.01$ $(0.08\pm 0.01)$}&\multirow{1}{*}{$0.41\pm 0.08$ $(0.39\pm 0.07)$} \\ \hline

\multirow{1}{*}{MPNetPath:NP (AC)}& \multirow{1}{*}{$0.03\pm 0.01$ $(0.03\pm 0.01)$} & \multirow{1}{*}{$0.06\pm 0.01$ $(0.06\pm 0.01)$}  &\multirow{1}{*}{ $0.08\pm 0.01$ $(0.08\pm 0.01)$}&\multirow{1}{*}{$0.53\pm 0.12$ $(0.42\pm 0.08)$} \\ \hline

\multirow{1}{*}{MPNetPath:NP (B)}& \multirow{1}{*}{$0.02\pm 0.00$ $(0.02\pm 0.00)$} & \multirow{1}{*}{$0.04\pm 0.00$ $(0.04\pm 0.01)$}  &\multirow{1}{*}{ $0.06\pm 0.01$ $(0.07\pm 0.01)$}&\multirow{1}{*}{$0.38\pm 0.04$ $(0.37\pm 0.02)$} \\ \hline

\multirow{1}{*}{MPNetPath:HP (B)}& \multirow{1}{*}{$0.04\pm 0.03$ $(0.07\pm 0.04)$} & \multirow{1}{*}{$0.13\pm 0.07$ $(0.14\pm 0.09)$}  &\multirow{1}{*}{ $0.09\pm 0.03$ $(0.12\pm 0.04)$}&\multirow{1}{*}{$0.42\pm 0.30$ $(0.41\pm 0.40)$} \\ \hline

\end{tabular}
\caption{Mean computation times with standard deviations are presented for MPNet (all variations), Informed-RRT* and BIT* on two test datasets, i.e., seen and unseen (shown inside brackets), in four different environments. In all cases, MPNet path planners, MPnetPath:NP (with neural replanning) and MPnetPath:HP (with hybrid replanning), and sampler MPNetSMP with underlying RRT*, trained with continual learning (C), active continual learning (AC) and offline batch learning (B), performed significantly better than classical planners such as Informed-RRT* and BIT* by an order of magnitude.} \label{tab1}
\end{table*}

\begin{figure*}[t]
    \centering
      \includegraphics[width=1.0\textwidth]{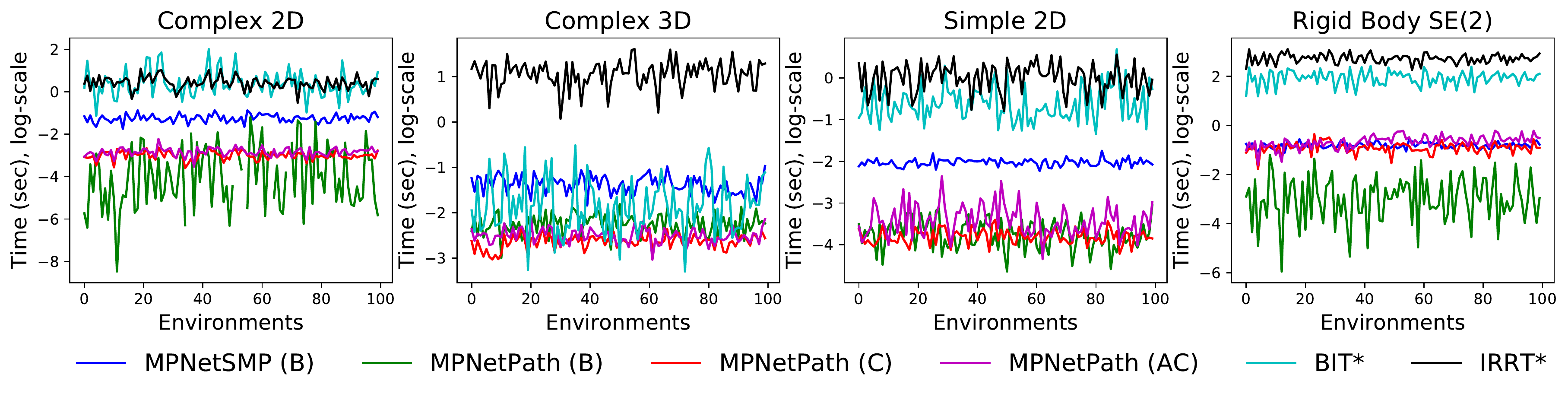}
    \caption{Mean computation time (log-scale) comparisons of MPNetPath (with neural replanning) and MPNetSMP (with underlying RRT*) trained with offline batch learning (B), continual learning (C) and active continual learning (AC) against Informed-RRT* (IRRT*) and BIT* on seen-$\mathcal{X}_\mathrm{obs}$ dataset that comprises 100 environments and each environment contains 200 planning problems.}\label{mpnet_graph1}
\end{figure*}


\begin{figure*}[t]
    \centering
      \includegraphics[width=1.0\textwidth]{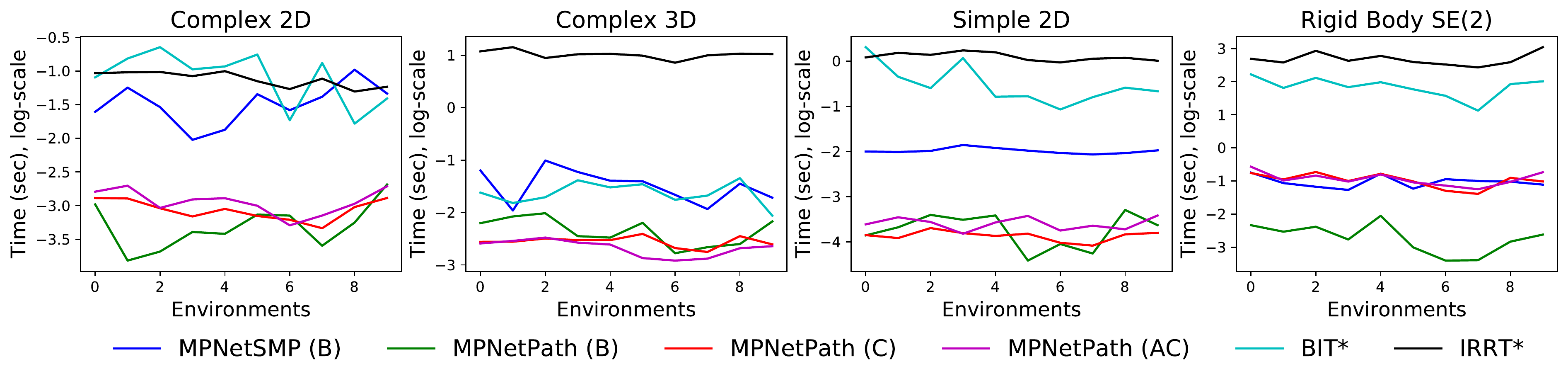}
    \caption{Mean computation time (log-scale) comparisons of MPNetPath (with neural replanning) and MPNetSMP (with underlying RRT*) against Informed-RRT* (IRRT*) and BIT* on unseen-$\mathcal{X}_\mathrm{obs}$ dataset that comprises unseen 10 environments and each environment contains 2000 planning problems. The training methods for MPNet includes offline batch learning (B), continual learning (C) and active continual learning (AC).}\label{mpnet_graph2}
\end{figure*}


\begin{table*}
\centering 
\begin{tabular}{|c|c|c|c|c|}\hline
\multirow{2}{*}{Methods}&\multicolumn{4}{c|}{Environments}\\\cline{2-5}
&\multicolumn{1}{c|}{Simple 2D }&\multicolumn{1}{c|}{Complex 2D}&\multicolumn{1}{c|}{Complex 3D}&\multicolumn{1}{c|}{Rigid-body-SE2}\\\hline

\multirow{1}{*}{MPNetPath:NP (C)}& \multirow{1}{*}{$93.33$ $(93.18)$} & \multirow{1}{*}{$83.44$ $(83.78)$}  &\multirow{1}{*}{ $89.88$ $(90.86)$}&\multirow{1}{*}{$83.770$ $(86.18)$}\\ \hline

\multirow{1}{*}{MPNetPath:NP (AC)}& \multirow{1}{*}{$96.70$ $(97.83)$} & \multirow{1}{*}{$84.36$ $(84.08)$}  &\multirow{1}{*}{ $96.60$ $(95.28)$}&\multirow{1}{*}{$87.08$ $(87.64)$} \\ \hline

\multirow{1}{*}{MPNetPath:NP (B)}& \multirow{1}{*}{$99.30$ $(98.30)$} & \multirow{1}{*}{$99.71$ $(98.80)$}  &\multirow{1}{*}{ $99.11$ $(97.76)$}&\multirow{1}{*}{$94.21$ $(95.18)$}  \\ \hline

\multirow{1}{*}{MPNetPath:HP (B)}& \multirow{1}{*}{$100.0$ $(100.0)$} & \multirow{1}{*}{$100.0$ $(100.0)$}  &\multirow{1}{*}{ $100.0$ $(100.0)$}&\multirow{1}{*}{$100.0$ $(100.0)$}  \\ \hline

\multirow{1}{*}{MPNetSMP}& \multirow{1}{*}{$100.0$ $(100.0)$} & \multirow{1}{*}{$100.0$ $(100.0)$}  &\multirow{1}{*}{ $100.0$ $(100.0)$}&\multirow{1}{*}{$100.0$ $(100.0)$}  \\ \hline

\end{tabular}
\caption{Success rates of all MPNet variants in the four environments on both test datasets, seen and unseen (shown inside brackets).} \label{tab2}
\end{table*}

\begin{figure}[h!]
    \centering
       \includegraphics[width=7.5cm]{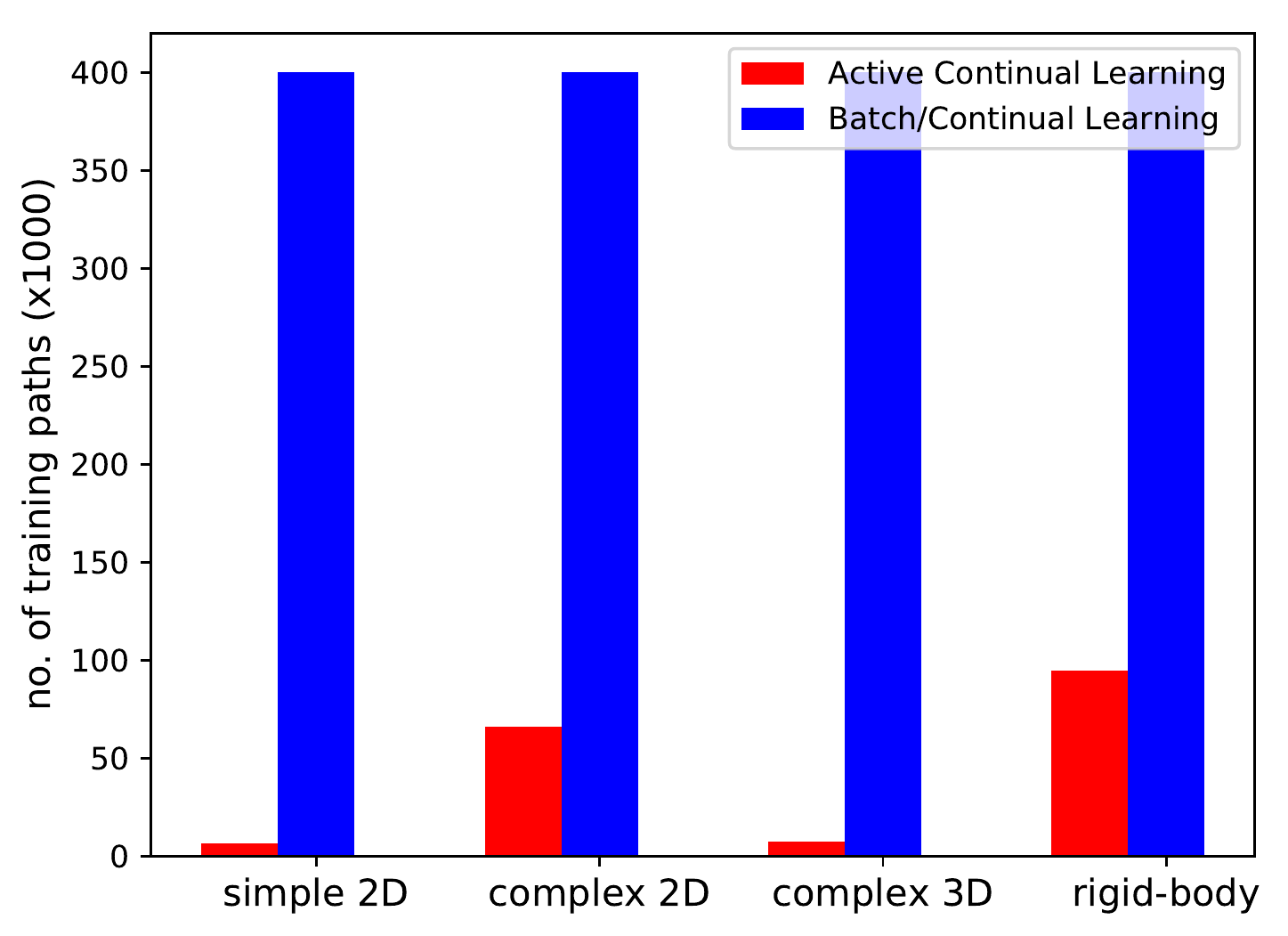}
    \caption{The number of training paths required by MPNet when trained with active continual learning as compared to traditional learning approaches. It can be seen that active continual learning improves data-efficiency and yet provides similar performance as conventional methods (Table I \& II).}\label{pathcount}
\end{figure}

\begin{figure*}[t]
    \centering
      \includegraphics[width=17.5cm]{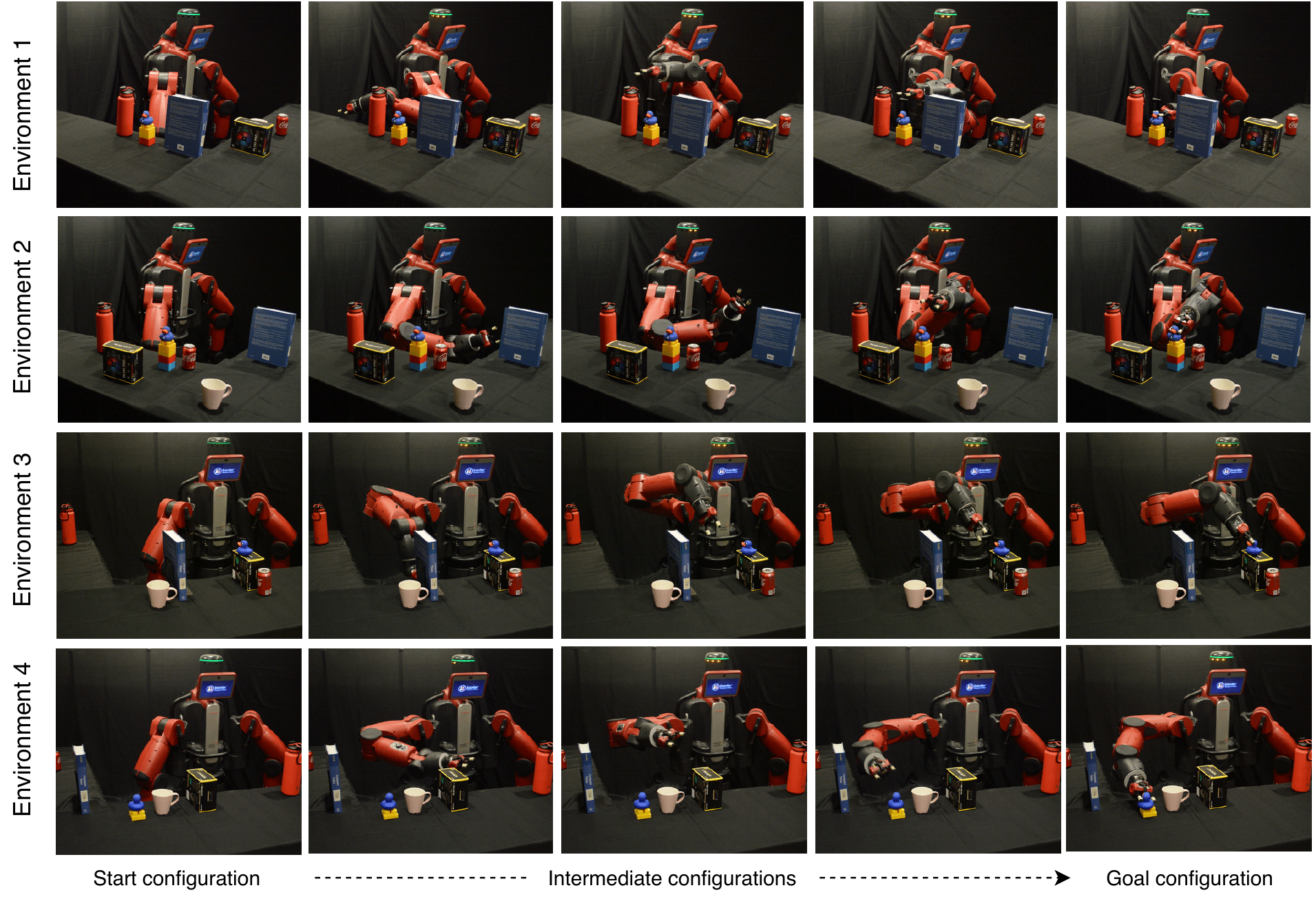}
    \caption{ We evaluated MPNetPath (with neural replanning) to plan motion for a Baxter robot in ten challenging and cluttered environments, out of which four are shown in this figure. The figures on the left-most and right-most indicate the robot at initial and goal configurations, respectively, whereas the blue duck shows the target positions. In these scenarios, MPNet took less than a second whereas BIT* took about 9 seconds to find solutions. In  these settings, we observed that BIT* solutions has almost 40\% higher path lengths than MPNet solutions.}\label{baxter1}
\end{figure*}
\section{Results}
In this section, we present results evaluating MPNet against state-of-the-art classical planners: RRT* \cite{karaman2011sampling}, Informed-RRT* \cite{gammell2014informed}, and BIT* \cite{gammell2015batch}. We use standard and efficient OMPL \cite{sucan2012the-open-motion-planning-library} implementations for classical planners. MPNet models were trained with the PyTorch Python API, exported using TorchScript \footnote{https://pytorch.org/tutorials/advanced/cpp\_export.html} so that they could be loaded in OMPL for execution. Furthermore, for MPNet, we present the results of MPNetPath and MPNetSMP trained with offline batch learning (B), continual learning (C), and active continual learning (AC). The MPNetPath:NP and MPNetPath:HP uses neural and hybrid replanning, respectively. The hybrid replanning exploits neural replanner for a fixed number of iterations $N_r$. The system used for experiments has 3.40GHz$\times$ 8 Intel Core i7 processor with 32 GB RAM and GeForce GTX 1080 GPU. 

\subsection{Training \& Testing Environments}
We consider problems requiring the planning of 2D/3D point-mass robot, rigid-body-SE2, rigid-body-SE3, and a 7DOF Baxter robot manipulator. We used encoder-decoder based unsupervised learning in the case of point-mass robots, and rigid-body-SE2. In Baxter and rigid-body-SE3 settings, we train Enet and Pnet together in an end-to-end setting.

In planning of 2D/3D point-mass and rigid-body-SE2, we train MPNet over one hundred workspaces with each containing four thousand trajectories, and it is evaluated on two test datasets, i.e., seen-$\mathcal{X}_\mathrm{obs}$ and unseen-$\mathcal{X}_\mathrm{obs}$. The seen-$\mathcal{X}_\mathrm{obs}$ comprises of one hundred workspaces seen by MPNet during the training but two hundred unseen start and goal pairs in each environment. The unseen-$\mathcal{X}_\mathrm{obs}$ consists of ten new workspaces, not seen by MPNet during training, with each containing two thousand start and goal pairs. For the 7DOF Baxter manipulator, our training dataset containing ten cluttered environments with each having nine hundred trajectories, and our test dataset contained the same ten workspaces as in training but one hundred new start and goal pairs for each scenario. In rigid-body-SE3 planning, we consider OMPL's \cite{sucan2012the-open-motion-planning-library} home-like environment, with training and testing dataset comprising two thousand trajectories and five-hundred unseen start and goal 3D poses of the rigid-body, respectively. For all planners presented in our experiments, we evaluate multiple trials on the given test dataset, and their mean performance metrics are reported.

\subsection{MPNet Comparison with its Expert Demonstrator (RRT*)}
We compare MPNet against its expert demonstrator RRT*. Fig. \ref{mpnetpath} shows the paths generated by MPNetPath (red), and its expert demonstrator RRT* (blue). The average computations times of MPNetPath and RRT* are denoted as $t_\mathrm{R}$ and $t_\mathrm{MP}$, respectively. The average Euclidean path cost of MPNetPath and RRT* solutions are denoted as $c_\mathrm{R}$ and $c_\mathrm{MP}$, respectively. It can be seen that MPNet finds paths of similar lengths as its expert demonstrator RRT* while retaining consistently low computational time. Furthermore, the computation times of RRT* are not only higher than MPNetPath computation time but also grows exponentially with the dimensionality of the planning problems. Overall, we observed that MPNetPath is at least $100\times$ faster than RRT* and finds paths that are within a 10\% range of RRT*'s paths cost.
  
Fig. \ref{mpnetsmp} presents informed trees generated by MPNetSMP with an underlying RRT* algorithm. We report average path computation times and Euclidean costs denoted as $t$ and $c$, respectively. The generated trees by MPNetSMP are in a subspace of given configuration space that most of the time contains path solutions. It should be noted that MPNetSMP paths are almost optimal and are observed to be better than MPNetPath and its expert demonstrator RRT* solutions. Moreover, our sampler not only finds near-optimal/optimal paths but also exhibit a consistent computation time of less than a second in all environments which is much lower than the computation time of RRT* algorithm. 

\subsection{MPNet Comparison with Advanced Motion Planners}
We further extend our comparative studies to evaluate MPNet against advanced classical planners such as Informed-RRT* 
\cite{gammell2014informed} and BIT* \cite{gammell2015batch}.  

Table I presents the comparison results over the four different scenarios, i.e., simple 2D (Fig. \ref{mpnetpath} (a-b)), complex 2D (Fig. \ref{mpnetpath} (c-d)), complex 3D (Fig. \ref{mpnetpath} (e-f)) and rigid-body-SE2 (Fig. \ref{mpnetpath} (g-h)). Each scenario comprises of two test datasets, seen-$\mathcal{X}_\mathrm{obs}$ and unseen-$\mathcal{X}_\mathrm{obs}$. We let Informed-RRT* and BIT* run until they find a solution of Euclidean cost within $5\%$ range of the cost of MPNetPath solution. We report mean computation times with standard deviation over five trials. In all planning problems, MPNetPath:NP (with neural replanning), MPNetPath:HP (with hybrid replanning), and MPNetSMP  exhibit a mean computation time of less than a second. The state-of-the-art classical planners, Informed-RRT* and BIT*, not only exhibit higher computation times than all versions of MPNet but, just-like RRT*, their computation times increase with the planning problem dimensionality. In simplest case (Fig. \ref{mpnetpath} (a-b)), MPNetPath:NP stand out to be atleast $80 \times$ and $30 \times$ faster than BIT* and Informed-RRT*, respectively. On the other hand, MPNetSMP provides about $8 \times$ and $5 \times$ computation speed improvements compared to BIT* and Informed-RRT*, respectively, in simple 2D environments. Furthermore, it can be observed that the speed gap between classical planner and MPNet (MPNetPath and MPNetSMP) keeps increasing with planning problem dimensions.   

Fig. \ref{mpnet_graph1} and Fig. \ref{mpnet_graph2} present the mean computation time, in log-scale, of all MPNet variants, Informed-RRT* (IRRT*), and BIT* on seen-$\mathcal{X}_\mathrm{obs}$ and unseen-$\mathcal{X}_\mathrm{obs}$ test datasets, respectively.  Note that MPNetPath trained with offline batch learning (B), continual learning (C), and active continual learning (AC) show similar computation times as can be seen by their plots in all planning problems. Furthermore, in Figs. \ref{mpnet_graph1}-\ref{mpnet_graph2}, it can be seen that MPNetPath and MPNetSMP computation times remain consistently less than one second in all planning problems irrespective of their dimensions. The computation times of IRRT* and BIT* are not only high but also lack consistency and exhibits high variations over different planning problems. Although the computation speed of MPNetSMP is slightly lower than that of MPNetPath, it performs better than all other methods in terms of path optimality for a given cost function.    

\subsection{MPNet Data Efficiency \& Performance Evaluation}
Fig. \ref{pathcount} shows the number of training paths consumed by all MPNet versions and Table II presents their mean success rates on the test datasets in four scenarios- simple 2D, complex 2D, complex 3D, and rigid-body-SE2. The success rate represents the percentage of planning problems solved by a planner in a given test dataset. The models trained with offline batch learning provides better performance in term of a success rate compared to continual/active-continual learning methods. However, in our experiments, the continual/active-continual learning frameworks required about ten training epochs compared to one hundred training epochs of the offline batch learning method. Furthermore, we observed that continual/active-continual learning and offline batch learning show similar success rates if allowed to run for the same number of training epochs.  The active continual learning is ultimately preferred as it requires less training data than traditional continual and offline batch learning while exhibiting considerable performance. 

\subsection{MPNet on 7DOF Baxter}
\begin{figure}[t]
    \centering
    \begin{subfigure}[b]{0.56\textwidth}
       \includegraphics[width=8.6cm]{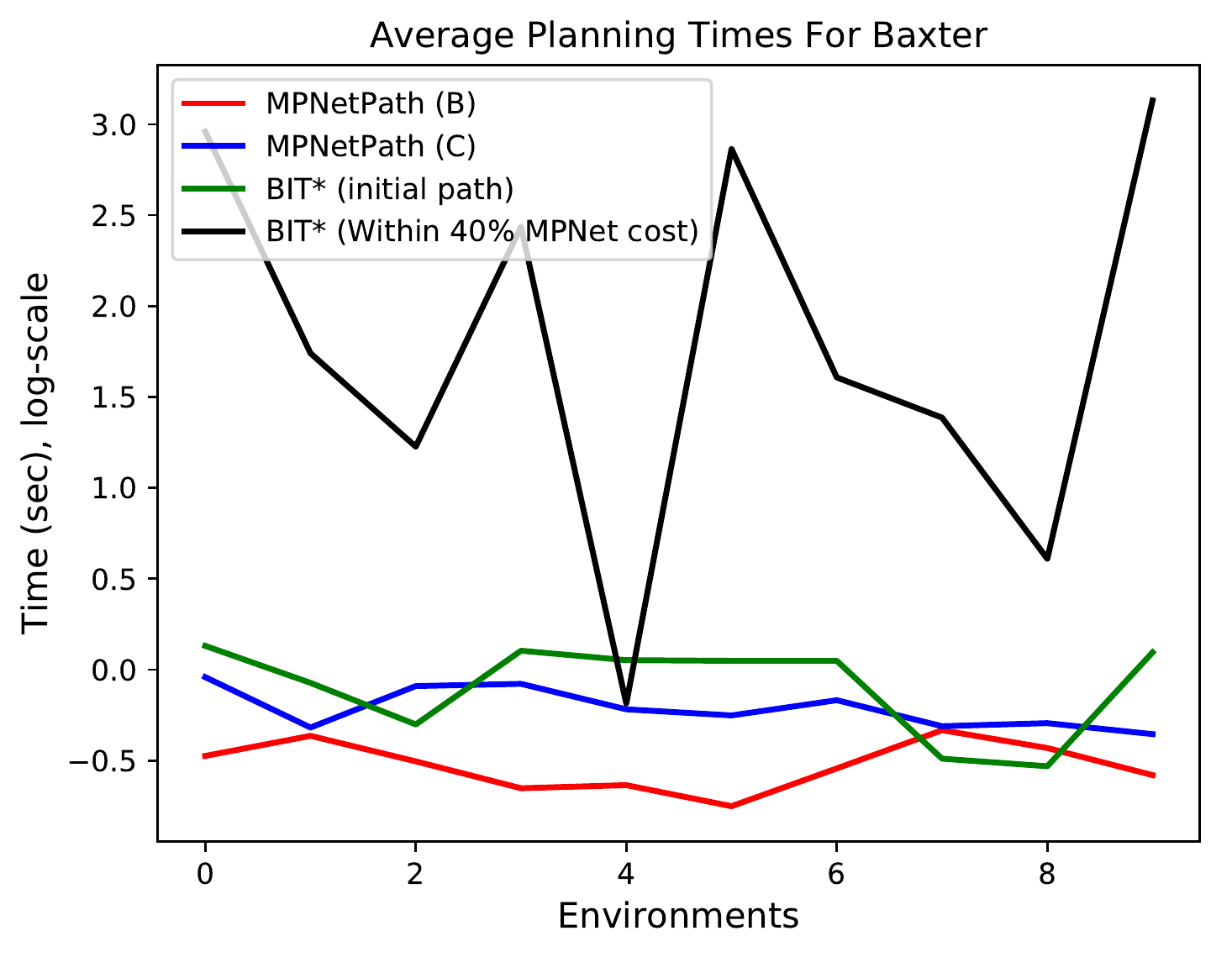}
    \end{subfigure}
        \begin{subfigure}[b]{0.56\textwidth}
       \includegraphics[width=8.6cm]{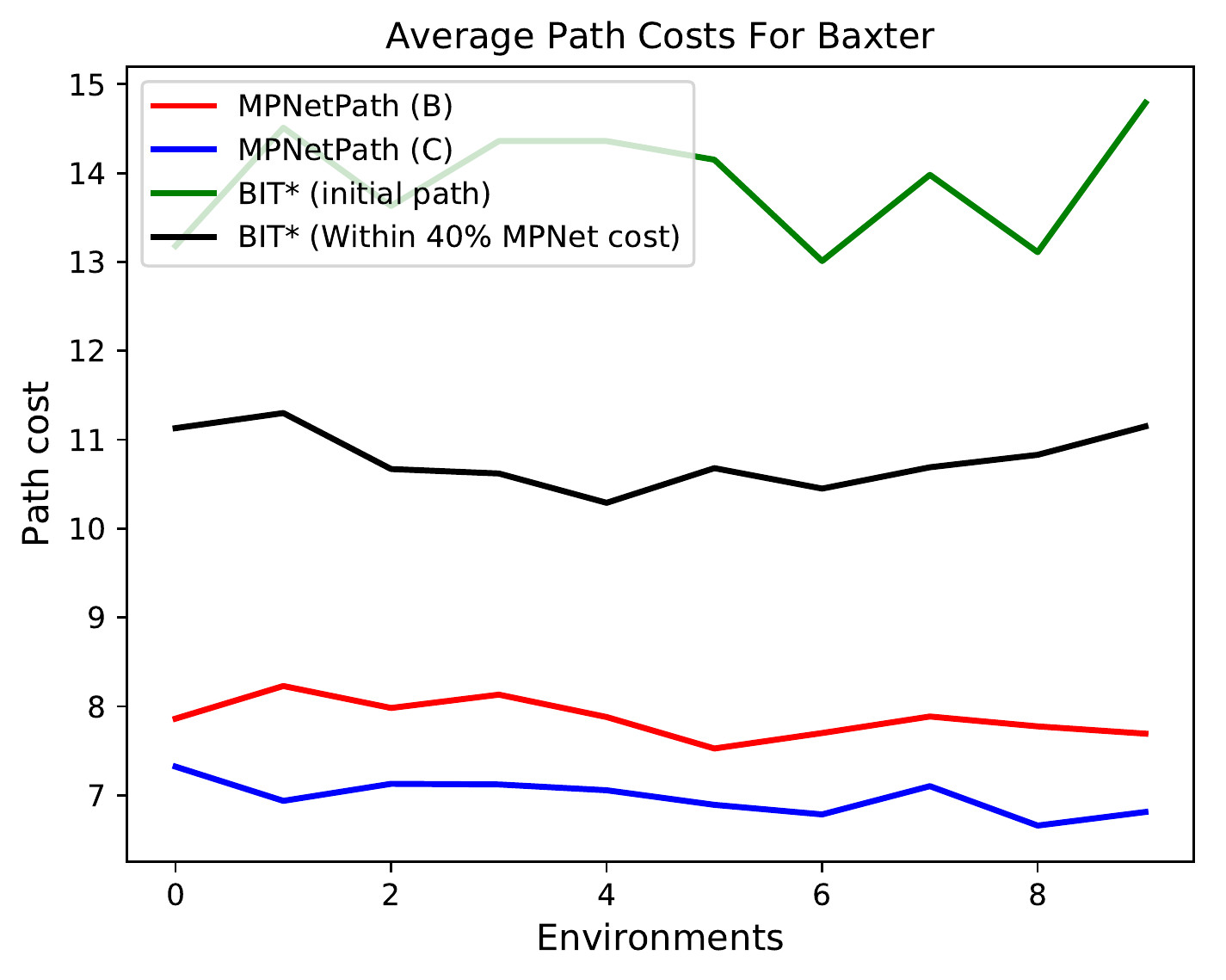}
    \end{subfigure}
    \caption{Computational time (log-scale) and path length comparison of MPNet (trained using batch (B) and continual (C) training) against BIT* over ten challenging environments that required path planning of a 7DOF Baxter. BIT* exhibits similar computation time for finding initial paths as MPNet, but the path lengths of BIT* solutions are significantly larger than path lengths of MPNet solutions.}\label{baxter4}
\vspace*{-0.2in} \end{figure}

Since BIT* performs better than Informed-RRT* in high dimension planning problems, we consider only BIT* as our classical planner in our further comparative analysis. We evaluate MPNet against BIT* for the motion planning of 7DOF Baxter robot in ten different environments where each environment comprised of one hundred new planning problems. Fig. \ref{baxter1} presents four out of ten environment settings. Each planning problem scenario contained an L-shape table of half the Baxter height with randomly placed five different real-world objects such as a book, bottle, or mug as shown in the figure. The planner's objective is to find a path from a given robot configuration to target configuration while avoiding any collisions with the environment, or self-collisions. Table III summarizes the results over the entire test dataset. MPNet on average find paths in less than a second with about 80\% success rate. BIT* also finds the initial path in less than a second with similar success rate as MPNet. However, the path costs of BIT* initial solution are significantly higher than the path cost of MPNet solutions. Furthermore, we let BIT* run to improve its initial path solutions so that the cost of the paths are not greater than 140\% of MPNet path costs. The results show that BIT* mostly fails to find solutions as optimal as MPNet solutions but demonstrate about 56\% success rate in finding solutions that are usually 40\% larger in lengths/costs than MPNet path costs.  Furthermore, the computation time of BIT* also increases significantly, requiring an average of about 9 seconds to plan such paths. Fig. \ref{baxter4} shows the mean computation time and path cost comparison of BIT* and MPNet over ten individual environments. It can be seen that MPNet computes paths in less than a second while giving incredibly optimized solutions and is much faster compared to BIT*.

\begin{figure*}[t]
    \centering
      \includegraphics[width=18cm]{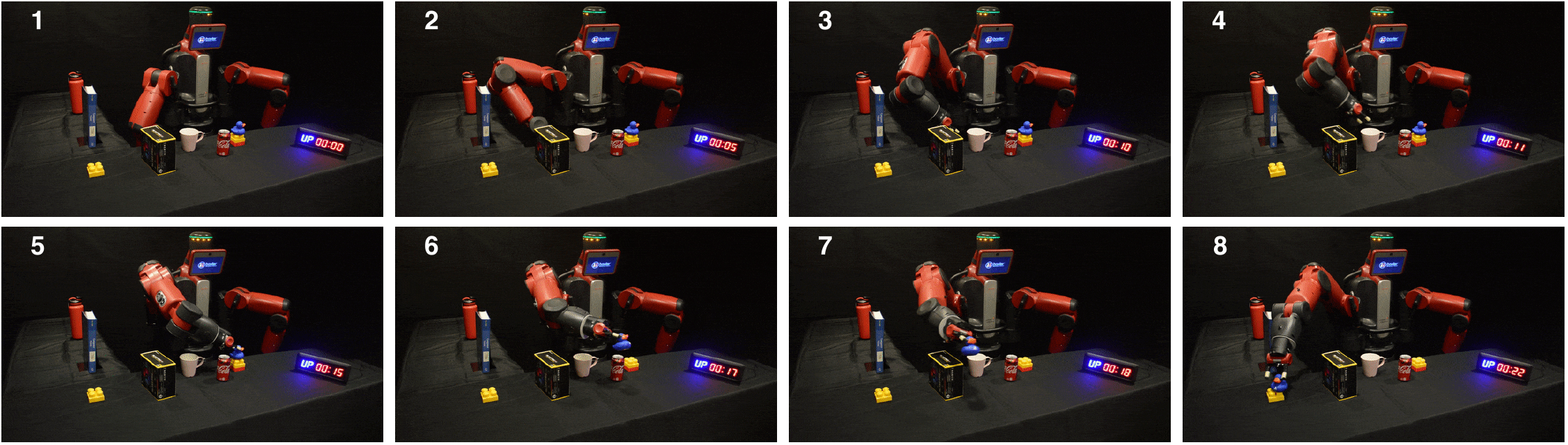}
    \caption{MPNetPath plans motion for a multi-target problem. The task is to pick up the blue object (duck), by planning a path from the initial robot position (Frame 1) to graspable object location (Frame 5), and move it to a new target (yellow block) (Frame 8). Note that the stopwatch indicates the execution time not the planning time. Furthermore, MPNet was never trained on any path connecting any two positions on the table-top, yet MPNet successfully planned the motion which indicates its ability to generalize to new planning problems.  In this scenario, MPNetPath (with neural replanning) computed the entire path plan in subsecond whereas BIT* took about few minutes to find a solution that is within $10\%$ cost of MPNet path solution.}\label{baxter2}
\end{figure*}

\begin{figure}[t]
    \centering
    \begin{subfigure}[b]{0.56\textwidth}
       \includegraphics[width=8.4cm]{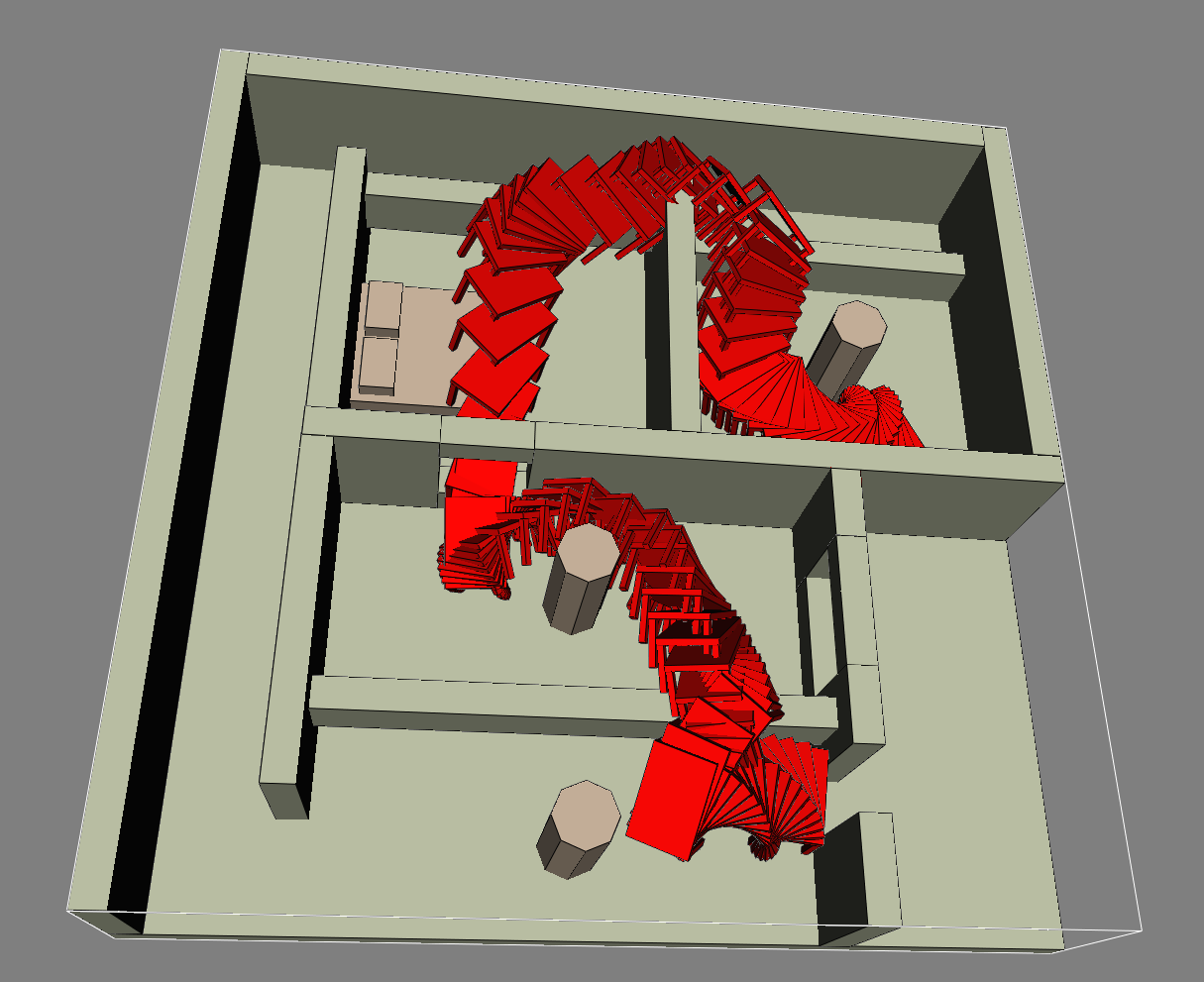}
    \end{subfigure}
        \begin{subfigure}[b]{0.56\textwidth}
       \includegraphics[width=8.4cm]{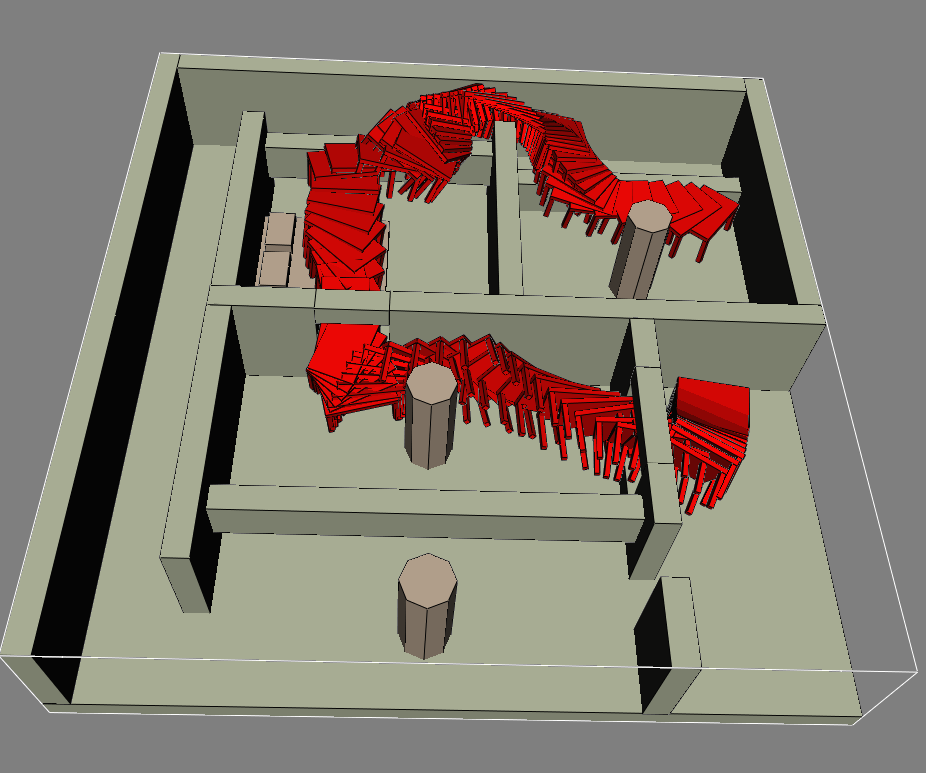}
    \end{subfigure}
    \caption{MPNet plans the motion of a rigid-body in SE(3) in a cluttered home-like environment with multiple narrow passages for randomly selected start and goal poses. }\label{se3}
\vspace*{-0.2in} \end{figure}

Fig. \ref{baxter2} shows a multi-goal planning problem for 7 DOF Baxter robot. The task is to reach and pick up the target object while avoiding any collision with the environment and transfer it to another target location, shown as yellow block. Note that in previous experiments we let BIT* run until it finds paths within $40\%$ of MPNet path costs. Now, we let BIT* run until it finds a path within $10\%$ cost of the cost of MPNet path solution. Our result shows that MPNetPath:NP plans the entire trajectory in less than a second whereas BIT* took about 3.01 minutes to find a path of similar cost as our solution. Note that MPNet is never trained on trajectories that connect two configurations on the table. However, thanks to MPNet ability to generalize that it can solve completely unseen problems in no time compared to classical planners that have to perform an exhaustive search before proposing any path solution. 
\begin{table*}
\centering 
\begin{tabular}{|c|c|c|c|}\hline
\multirow{2}{*}{Methods}&\multicolumn{3}{c|}{Baxter}\\\cline{2-4}
&\multicolumn{1}{c|}{Time}&\multicolumn{1}{c|}{Path cost}&\multicolumn{1}{c|}{Success rate$(\%)$}\\\cline{1-4}
BIT* (Initial Path)&$0.94 \pm 0.20$&$13.91 \pm 0.60$&$83.0$\\\hline
BIT* ($\pm40\%$ MPNet cost)&$9.20 \pm 7.61$&$10.78 \pm 0.31$&$56.0$\\\hline
MPNetPath (C)&$0.81 \pm 0.08$&$6.98 \pm 0.18$&$78.6$\\\hline
MPNetPath (B)&$0.59 \pm 0.08$&$7.86 \pm 0.20$&$87.8$\\\hline
\end{tabular}
\caption{Computation time, path cost/length, and success rate comparison of MPNetPath (with neural replanning) trained with offline batch (B) and continual (C) learning against BIT* on Baxter test dataset. In case of BIT*, the times for finding the first path and further optimizing it to a path cost within 40\% range of MPNet's path cost are reported.  Overall, MPNet exhibits superior performance than BIT* in terms of path costs, mean computation times, and success rates.}\label{tab3}
\vspace*{-0.2in}\end{table*}
{\subsection{MPNet planning in SE (3) }
To further extend our comparison of MPNet and BIT*, we also consider planning in SE (3) for a rigid-body in a cluttered home-like environment with multiple narrow passages \cite{sucan2012the-open-motion-planning-library}, as shown in the Fig. \ref{se3}. Our test dataset comprised 500 randomly sampled start and goal poses (not seen by MPNet during training), and the Fig. \ref{se3} shows two of those test cases. In these settings, MPNet exhibits a success rate of about $85\%$. The mean computation time to find an initial path solution for MPNet (B), MPNet (C), and BIT* were $0.96$, $1.61$, and $2.84$ seconds, respectively. We also observed that MPNet paths' cost were significantly lower than the cost of BIT* solutions. The mean path costs of MPNet (B), MPNet (C), and BIT* were $457.80$, $512.73$, and $836.85$, respectively. Furthermore, we noticed that BIT* takes up to several minutes to find a path of similar cost as MPNet's solution. }

\section{Discussion}
This section presents a discussion on various attributes of MPNet, sample selection methods for continual learning, and a brief analysis of its completeness, optimality, and computational guarantees. 
\subsection{Stochastic Neural Planning \& Replanning}
In MPNet, we apply Dropout \cite{srivastava2014dropout} to almost every layer of the planning network, and it remains active during path planning. The Dropout randomly selects the neurons from its preceding layer with a probability $p \in [0,1]$ and masks them so that they do not affect the model decisions. Therefore, at any planning step, Pnet would have a probabilistically different model structure that results in stochastic behavior. 

Yarin and Zubin \cite{gal2016dropout} demonstrate the application of Dropout to model uncertainty in the neural networks. We show that the Dropout can also be used in learning-based motion planners to approximate the subspace of a given configuration space that potentially contains several path solutions to a given planning problem. For instance, it is evident from the trees generated by MPNetSMP (Fig. \ref{mpnetsmp}) that our model approximates the subspace containing path solutions including the optimal path. 

The perturbations in Pnet's output, thanks to Dropout, also play a crucial role during our neural replanning (Algorithm \ref{algo:RNP}). In neural replanning, MPNet takes its own global path and uses its stochastic neural planner (Algorithm \ref{alg:mpnetpath}) to replan a motion between any beacon states. Replanning is a crucial component of our planning algorithm. Although the global plan comprises collision-free nodes, the straight-line connection between those nodes might not be collision-free (Fig. \ref{mpnet3} (b)). We observed that without replanning, MPNet exhibits a low success rate of about $60-70\%$. Since global planning decomposes a given problem into sub-problems, the replanning between beacon nodes to solve sub-problems through our stochastic neural planner helps our method to recover from any failures leading to a high success rate in complex, cluttered environments.
\subsection{Sample Selection Strategies}
\begin{figure}[h!]
    \centering
    \begin{subfigure}[b]{0.56\textwidth}
       \includegraphics[width=8.5cm]{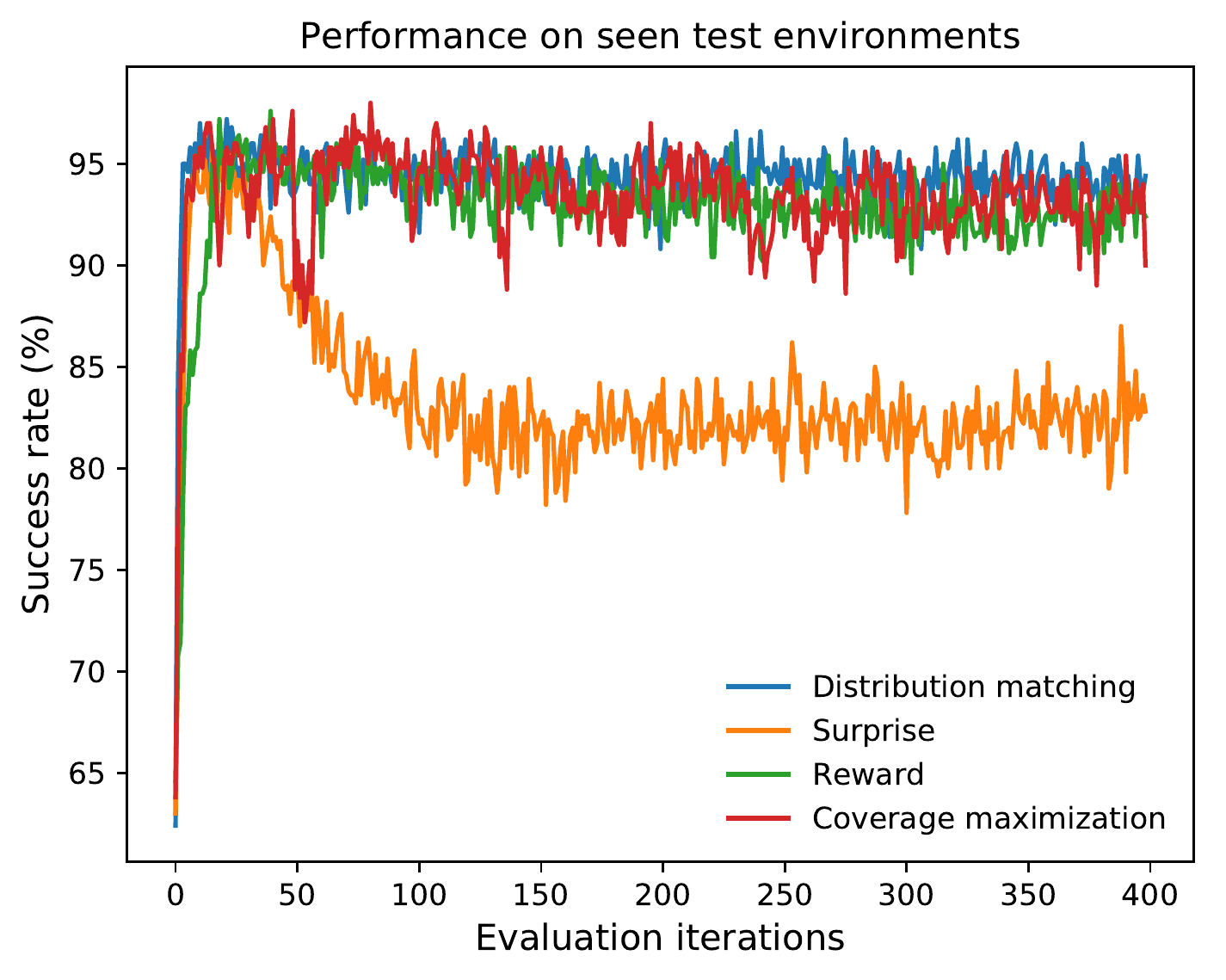}
    \end{subfigure}
        \begin{subfigure}[b]{0.56\textwidth}
       \includegraphics[width=8.5cm]{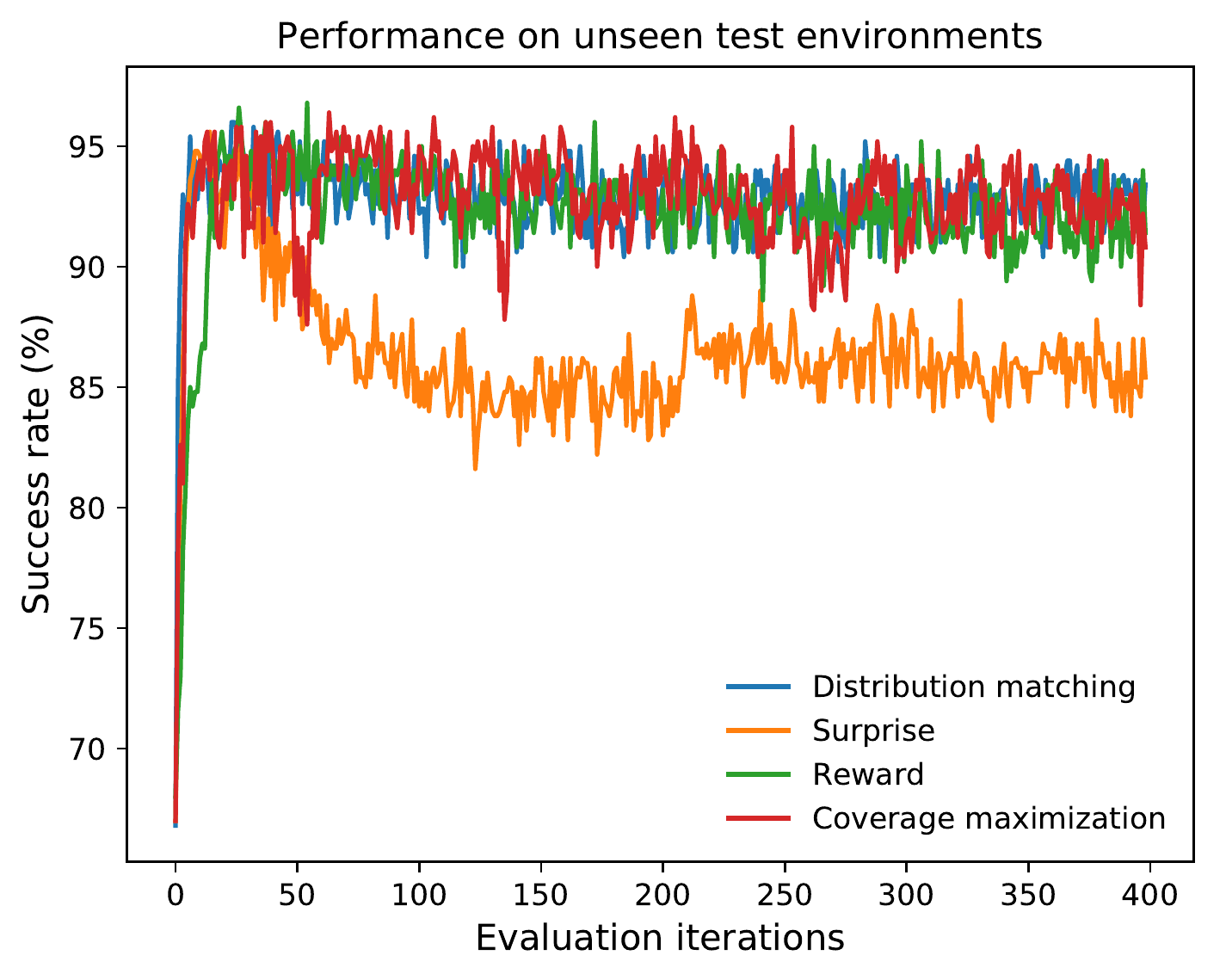}
    \end{subfigure}
    \caption{Impact of sample selection methods for episode memory on the performance of continual learning for MPNet in simple 2D test datasets seen-$\mathcal{X}_\mathrm{obs}$ and unseen-$\mathcal{X}_\mathrm{obs}$.}\label{em}
\end{figure}

In continual learning, we maintain an episodic memory of a subset of past training data to mitigate catastrophic forgetting when training MPNet on new demonstrations. Therefore, it is crucial to populate the episodic memory with examples that are important for learning a generalizable model from streaming data. We compare four MPNet models trained on a simple 2D environment with four different sample selection strategies for the episodic memory. The four selection metrics include surprise, reward, coverage maximization, and global distribution matching. The surprise and reward metrics give higher and lower priority to examples with large prediction errors, respectively. The coverage maximization maintains a memory of $k$-nearest neighbors. The global distribution matching, as described earlier, uses random selection techniques to approximate the global distribution from streaming data. We evaluate four MPNet models on two test datasets, seen-$\mathcal{X}_\mathrm{obs}$ and unseen-$\mathcal{X}_\mathrm{obs}$, from simple 2D environment, and their mean success rates are reported  in Fig. \ref{em}. It can be seen that global distribution matching exhibits slightly better performance than reward and coverage maximization metrics. The surprise metric gives poor performance because satisfying the constraint in Equations \ref{gem3} becomes nearly impossible when the episodic memory is populated with examples that have large prediction losses. Since global distribution matching provides the best performance overall, we have used it as our sample selection strategy for the episodic memory in the continual learning settings. 
\subsection{Batch offline \& Active Continual Learning}
In this section, we briefly highlight the merits of our training approaches. A batch offline learning is preferred when plenty of data is available for offline training, especially in cases where existing planners can be used to generate data. However, in cases where data is expensive to make, an active continual learning approach is preferable as it asks for demonstration only when needed leading to data-efficient training of our models. For instance, in semi-autonomous driving, the planning problems would usually come in streams, and our neural planner will only ask for a human demonstration when needed. Also, to realize classical life-long learning one can combine both batch offline and active continual learning where models are pre-trained with the former and further refined with the latter whenever needed.


\subsection{Completeness}
In this section, we discuss the completeness guarantees of MPNet, formally proposed as follow:
\\\\
\textbf{Proposition 1 (Feasible Path Planning)}\textit{ Given a planning problem $\{c_\mathrm{init},c_\mathrm{goal},x_\mathrm{obs}\}$, and a collision-checker, MPNet finds a path solution $\sigma:[0,T]$, if one exists, such that $\sigma_0=c_\mathrm{init}$, $\sigma_T\in c_\mathrm{goal}$, and $ \sigma \subset \mathcal{C}_\mathrm{free}$. }\\ \\
Proposition 1 implies that for a given planning problem, MPNet will eventually find a feasible path, if one exists. We show that the worst-case completeness guarantees of our approach rely on the underlying SMP for both path planning (MPNetPath) and informed neural sampling (MPNetSMP). In our experiments, we use RRT* as our oracle SMP that exhibits \textit{probabilistic completeness}.

The \textit{probabilistic completeness} is described in Definition 1 based on the following notations. Let $\mathcal{T}^{AL}_n$ be a connected tree in obstacle-free space, comprising of $n$ vertices, generated by an algorithm $AL$. We also assume $\mathcal{T}^{AL}_n$ always originates from the robot's initial configuration $c_\mathrm{init}$. An algorithm is probabilistically complete if it satisfies the following definition.
\\\\ 
\textbf{Definition 1 (Probabilistic Completeness)}\textit{ Given a planning problem $\{c_\mathrm{init},c_\mathrm{goal}, \mathcal{X}_\mathrm{obs}\}$ for which there exists a solution, an algorithm AL with a tree $\mathcal{T}^{AL}_n$ that originates at $c_\mathrm{init}$ is considered probabilistically complete iff $\lim_{n \rightarrow \infty}\mathbb{P}(\mathcal{T}^{AL}_n \cap c_\mathrm{goal} \neq \varnothing)=1$.}\\

RRT* algorithm exhibits \textit{probabilistic completeness} as it builds a connected tree originating from initial robot configuration and randomly exploring the entire space until it finds a goal configuration. Therefore, as the number of samples in the RRT* approach to infinity the probability of finding a path solution, if one exists, gets to one, i.e.,
\begin{equation}
\lim_{n \rightarrow \infty}\mathbb{P}(\mathcal{T}^{RRT^*}_n \cap c_\mathrm{goal} \neq \varnothing)=1
\end{equation}

In remainder of the section, we present a brief analysis showing that MPNetPath and MPNetSMP also exhibit \textit{probabilistic completeness} like their underlying RRT*.\\
\subsubsection{Probabilistic completeness of MPNetPath}
To justify MPNetPath worst-case guarantees, we introduce the following assumptions that are crucial to validate a planning problem.\\\\
\textbf{Assumption 1 (Feasibility of States to Connect)}\textit{ The given start and goal pairs $(c_\mathrm{init}, c_\mathrm{goal})$, for which an oracle-planner is called to find a path, lie entirely in obstacle-free space, i.e., $c_\mathrm{init} \in \mathcal{C}_\mathrm{free}$ and $c_\mathrm{goal} \subset \mathcal{C}_\mathrm{free}$.}\\ \\
\textbf{Assumption 2 (Existence of a Path Solution)}\textit{ Under Assumption 1, for a given problem $(c_\mathrm{init}, c_\mathrm{goal}, \mathcal{X}_\mathrm{obs})$, there exists at least one feasible path $\sigma$ that connects $c_\mathrm{init}$ and $c_\mathrm{goal}$, and $\sigma\not \subset \mathcal{X}_\mathrm{obs}$.}\\ \\
Based on Definition 1 and Assumptions 1-2, we propose in Theorem 1, with a sketch of proof, that MPNetPath also exhibits \textit{probabilistic completeness}, just like its underlying oracle planner RRT*.\\\\
\textbf{Theorem 1 (Probabilistic Completeness of MPNetPath)}\textit{ If Assumptions 1-2 hold, for a given planning problem $\{c_\mathrm{init}, c_\mathrm{goal},\mathcal{X}_\mathrm{obs}\}$, and an oracle RRT* planner, MPNetPath will find a path solution, if one exists, with a probability of one as the underlying RRT* is allowed to run till infinity if needed.}\\ \\
\textit{Sketch of Proof:}
Assumption 1 puts a condition that the start and goal states in the given planning problem lie in the obstacle-free space. Assumption 2 requires that there exist at least one collision-free trajectory for the given planning problem. During planning, MPNetPath first finds a coarse solution $\sigma$ that might have beacon (non-connectable consecutive) states (see Fig. \ref{mpnet3}). Our algorithm connects those beacon states through replanning. Assumption 1 holds for the beacon states as each generated state of MPNetPath is evaluated by an oracle collision-checker before making it a part of $\sigma$. In replanning, we perform neural replanning for a fixed number of trials to further refine $\sigma$, and if that fails to conclude a solution, we connect any remaining beacon states of the refined $\sigma$ by RRT*. Hence, if Assumption 1-2 hold for beacon states, MPNetPath inherits the convergence guarantees of the underlying planner which in our case is RRT*. Therefore, MPNetPath with an underlying RRT* oracle planner exhibits probabilistic completeness. \\
\subsubsection{Probabilistic completeness of MPNetSMP}
MPNetSMP generates samples for SMPs such as RRT*. Our method performs exploitation and exploration to build an informed tree. It begins with exploitation by sampling a subspace that potentially contains a solution for a limited time and switches to exploration via uniform sampling to cover the entire configuration space. Therefore, like RRT*, the tree of MPNetSMP is fully connected, originates from the initial robot configuration, and eventually expands to explore the entire space. Hence, the probability that MPNetSMP tree will find a goal configuration, if possible, approaches one as the samples in the tree reaches a large number, i.e., 
\begin{equation}
\lim_{n \rightarrow \infty}\mathbb{P}(\mathcal{T}^{\mathrm{MPNetSMP}}_n \cap c_\mathrm{goal} \neq \varnothing)=1
\end{equation}

\subsection{Optimality}
MPNet learns through imitating the expert demonstrations which could come from either an oracle motion planner or a human demonstrator. In our experiments, we use RRT* for generating expert trajectories, hybrid replanning in MPNetPath, and as a baseline SMP for our MPNetSMP framework. Optimal SMPs such as RRT* exhibits asymptotic optimality, i.e., a probability of finding an optimal path, if one exists, approaches one as the number of random samples in the tree approaches infinity (for proof, refer to \cite{karaman2011sampling}). RRT* gets weak-optimality guarantees from its rewiring heuristic. The rewiring incrementally updates the tree connections such that over the time, each node in the tree would be connected to a branch that ensures lowest cost path to the root state (initial robot configuration) of the tree. 

In our experiments, we show that MPNetPath imitates the optimality of its expert demonstrator. In Fig. \ref{mpnetpath}, it can be seen that MPNet path solutions are of similar Euclidean costs as its expert demonstrator RRT* path solutions. Therefore, the quality of the MPNet paths relies on the quality of its training data. 

In the case of MPNetSMP, there exists optimality guarantees depending on the underlying SMP. Since we use RRT* as a baseline SMP for MPNetSMP, we formally propose in Proposition 2 that our method exhibits asymptotic optimality. 
\\ \\ 
\textbf{Proposition 2 (Optimal Path Planning)}\textit{ Given a planning problem $\{c_\mathrm{init},c_\mathrm{goal},x_\mathrm{obs}\}$, a collision-checker, and a cost function $J(\cdot)$, MPNet can adaptively sample the configuration space for an asymptotic optimal SMP such that the probability of finding an optimal path solution, if one exists, w.r.t. $J(\cdot)$ approaches one as the number of samples in the graph approaches infinity.}\\ \\
In our sampling heuristic, MPNetSMP generates informed samples for a fixed number of iterations and performs uniform random sampling afterward. Therefore, RRT* tree formed by the samples of MPNetSMP will explore the entire C-space and will be rewired incrementally to ensure asymptotic optimality. Since MPNetSMP only performs intelligent sampling and does not modify the internal machinery of RRT*, the optimality proofs will exactly be the same as provided in \cite{karaman2011sampling}. Hence, provided that MPNetSMP performs exploitation for fixed steps and pure exploration afterward, and the underlying SMP is RRT*, the optimality of our method is asymptotically guaranteed .  

\subsection{Computational Complexity}
In this section, we present the computational complexity of our method. Both MPnetPath and MPNetSMP take the output of Enet and Pnet, which is only a forward-pass through a neural network. The forward pass through a neural network is known to have a constant complexity since it is a matrix multiplication of a fixed size input vector with network weights to generate a fixed size output vector. Since MPNetSMP produces samples by merely forward passing a planning problem through Enet and Pnet, it does not add any complexity to the underlying SMP. Therefore, the computational complexity of MPNetSMP with underlying RRT* is the same as RRT* complexity, i.e., $O(n\log n)$, where $n$ is the number of samples \cite{karaman2011sampling}. 

On the other hand, MPNetPath has a crucial replanning component that uses an oracle planner in the cases where neural replanning fails to determine an end-to-end collision-free path. The oracle planner in the presented experiments is RRT*. Therefore, in the worst-case, MPNetPath operates with complexity of $O(n\log n)$. However, we show empirically that MPNetPath succeeds, without any oracle planner, for up to 90\% of the cases in a challenging test dataset. For the remaining 10\% cases, oracle planner is executed only for a small segment of the overall path (see Algorithm \ref{alg:mpnetpath}). Hence, even for hard planning problems where MPNetPath fails, our method reduces a complicated problem into a simpler problem, thanks to its divide-and-conquer approach, for the underlying oracle planner. Thus, the oracle planner operates at a complexity which is practically much lower than its worst-case theoretical complexity, as is also evident from the results of MPNetPath with hybrid replanning (HB) in Table I. 
 
\section{Conclusions and Future work }
We present a learning-based approach to motion planning using deep neural networks. For a given planning problem, our method is capable of i) finding collision-free near-optimal paths; ii) generating samples for sampling-based motion planners in a subspace of a given configuration space that most likely contains solutions including the optimal path. We also present the active continual learning strategy to train our models with a significant improvement in training data-efficiency compared to naive training approaches. Our experimentation shows that our neural motion planner consistently finds collision-free paths in less than a second for the problems where other planners may take up to several minutes. 

In our future works, one of our primary objectives is to tackle the environment encoding problem for motion planning. Environment encoding is one of the critical challenges in real-world robotics problems. Current perception approaches consider encoding of individual objects rather than the entire scene that retains the inter-object relational geometry which is crucial for motion planning. Henceforth, we aim to address the problem of motion planning oriented environments' encoding from raw point-cloud data. Another future objective is to learn all fundamental modules required for motion planning that not only include the motion planner but also the collision checker and cost function. Recently, L2RRT \cite{ichter2019robot} is proposed that learns the motion planner and collision checker in latent space. However, providing worst-case theoretical guarantees is notoriously hard for planning in latent spaces. Therefore, a modular approach is necessary that learns decision-making directly in robot configuration space so that it can be combined with any classical motion planner to provide theoretical guarantees. Furthermore, learning a cost function, using inverse reinforcement learning \cite{qureshi2018adversarial}, is also crucial for kinodynamic and constraint motion planning problems where defining a cost function satisfying all constraints is challenging. 

\bibliographystyle{IEEEtran}
\bibliography{references}
\nocite{*}

\end{document}